\documentclass[letterpaper]{article} 
\usepackage{aaai2027_arxiv}  
\usepackage[hyphens]{url}  
\usepackage{graphicx} 
\usepackage{natbib}  
\usepackage{caption} 
\usepackage{algorithm}
\usepackage{algorithmic}

\usepackage[dvipsnames]{xcolor}

\definecolor{Black}{rgb}{0.0, 0.0, 0.0}

\definecolor{DarkGreen}{rgb}{0.10, 0.55, 0.10}
\definecolor{DeepSkyBlue3}{rgb}{0.0, 0.686, 0.843}
\definecolor{DarkTurquoise}{rgb}{0.0, 0.843, 0.843}
\definecolor{Cyan3}{rgb}{0.0, 0.843, 0.686}
\definecolor{LightSeaGreen}{rgb}{0.0, 0.686, 0.686}

\definecolor{RoyalBlue}{rgb}{0.20, 0.60, 0.86}
\definecolor{DeepSkyBlue}{rgb}{0.0, 0.686, 1.}

\definecolor{DodgerBlue}{rgb}{0.0, 0.529, 1.}
\definecolor{DodgerBlue2}{rgb}{0.0, 0.3725, 1.}
\definecolor{DodgerBlue3}{rgb}{0.0, 0.3725, 0.843}
\definecolor{DarkCyan}{rgb}{0.0, 0.54, 0.54}

\definecolor{Gray}{gray}{0.9}
\definecolor{ChromeYellow}{rgb}{1.0, 0.65, 0.0}

\definecolor{Gold}{rgb}{1.0, 0.843, 0.0}

\definecolor{Crimson}{rgb}{0.86, 0.08, 0.24}
\definecolor{IndianRed}{rgb}{1.0, 0.373, 0.529}

\definecolor{SunsetOrange}{rgb}{0.98, 0.37, 0.33}
\definecolor{DarkOrange}{rgb}{1.0, 0.529, 0.}
\ifx \submission \undefined

\else

\fi

\usepackage{xcolor}
\usepackage{amsmath}
\usepackage{amssymb}
\usepackage{booktabs}
\usepackage[table]{xcolor}
\usepackage{multirow}
\usepackage{array}
\usepackage{graphicx}
\usepackage{placeins}

\definecolor{OursRow}{RGB}{235,245,255}
\definecolor{AvgCol}{RGB}{242,242,242}
\newcolumntype{A}{>{\columncolor{AvgCol}}c}
\newcommand{\bitslabel}[1]{\makebox[0.75em][c]{\rotatebox[origin=c]{90}{\textbf{#1}}}}
\newlength{\TableWidth}

\usepackage{newfloat}
\usepackage{listings}
\DeclareCaptionStyle{ruled}{labelfont=normalfont,labelsep=colon,strut=off} 
\floatstyle{ruled}
\newfloat{listing}{tb}{lst}{}
\floatname{listing}{Listing}

\usepackage{booktabs}

\title{ReRound: \underline{Re}constructive \underline{Round}ing to Resolve Midpoint Ambiguity \\ in Calibration-Free LLM Quantization}

\author{
    He-Yen Hsieh, H. T. Kung
}
\affiliations{
    Harvard University

}

\begin{document}

\maketitle

\begin{abstract}
ReRound (Reconstructive Rounding) is a post-training quantization method that
addresses the midpoint ambiguity inherent in standard round-to-nearest (RTN) schemes
when quantizing weights near the centers of quantization intervals.

Starting from a pretrained LLM, ReRound trains a conditional diffusion model to
produce continuous reconstructions of low-bit weights for the LLM. These reconstructed weights act as a
guidance signal to disambiguate the rounding direction of weights located close to
interval midpoints.

To integrate this reconstruction-guided rounding with conventional RTN, ReRound
introduces a tolerance metric measuring how far the quantized weight (not the final quantized integer) is away from the midpoint: quantized weights within a tolerance region around
midpoints are quantized using diffusion-based reconstructions, whereas weights closer
to quantization boundaries are quantized with RTN.
By sweeping the tolerance parameter, ReRound generates multiple candidate quantized integer
weight matrices and selects the de-quantized weight matrix candidate whose leading singular values most closely match
those of the original full-precision weights. This selected candidate determines the tolerance parameter ReRound uses.

ReRound is particularly effective for smaller LLMs. Across a range of such models, it
consistently outperforms standard RTN for 3-bit and 4-bit weight quantization. ReRound achieves superior accuracy compared to an extensive set of calibration-free methods, remains competitive with calibration-dependent approaches, and operates entirely offline,
introducing no additional overhead during low-bit inference.

The ReRound strategy represents a new approach for low-bit quantization. The method applies to AI models beyond LLMs. This paper focuses on its applications to small LLMs. Code is available at
\textcolor[HTML]{0F9ED5}{\url{https://github.com/louisYen/ReRound}}.
\end{abstract}

\begin{figure}[!t]
\centering
\includegraphics[width=0.99\columnwidth]{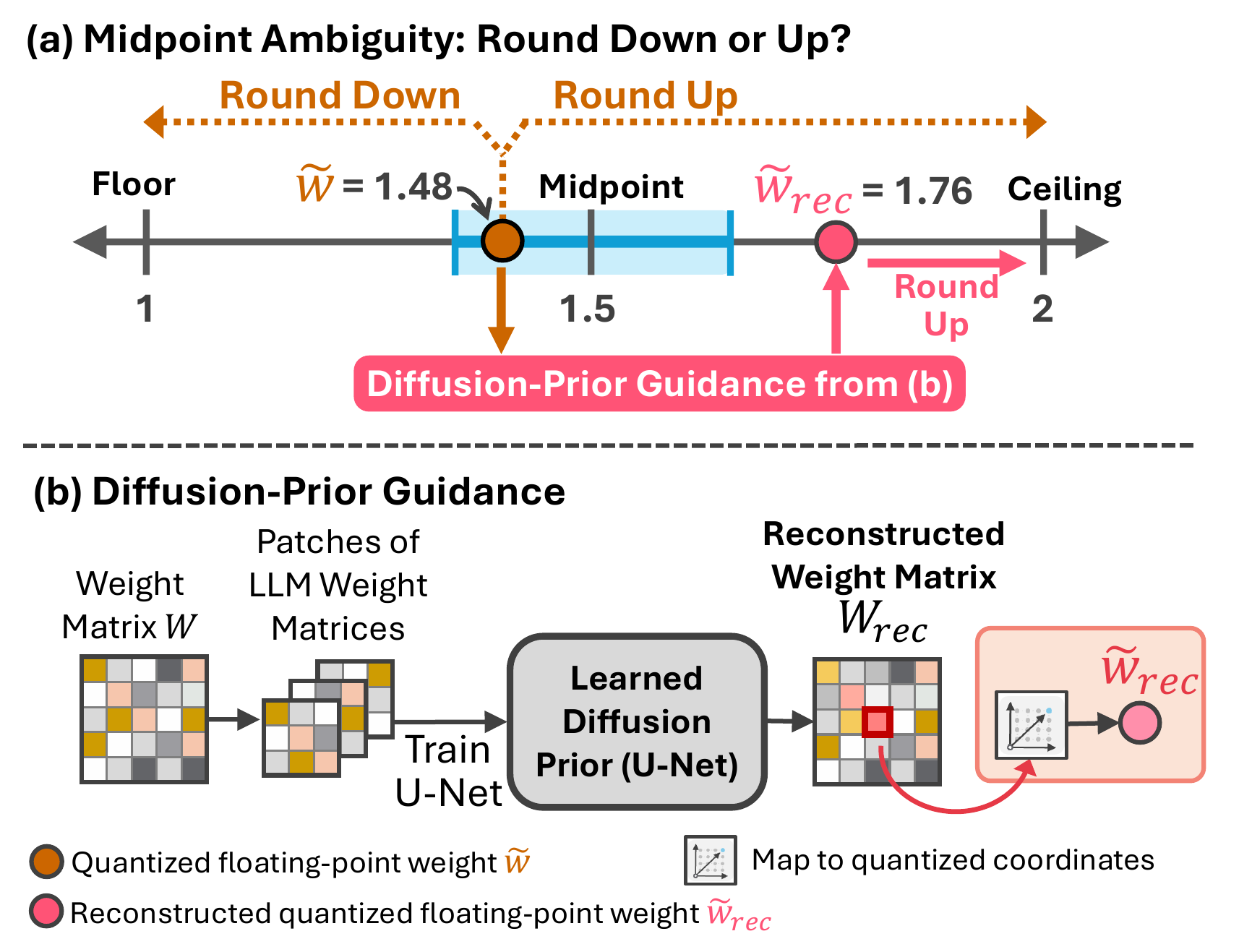}
\caption{
\textbf{ReRound resolves ambiguous rounding with a learned diffusion prior.}
\textbf{(a)}
The quantized floating-point weight $\widetilde{w}=1.48$ is in the
midpoint ambiguity region (\textcolor[HTML]{0F9ED5}{blue interval}) in the
sense that $\widetilde{w}$ is sufficiently near the midpoint $1.5$.
As a result, there is low confidence on whether $\widetilde{w}$ should be
rounded down to the floor $1$ or up to the ceiling $2$.
But, under RTN, because $\widetilde{w}$ is slightly closer to the floor
$1$, in this case it will always be rounded to the floor.
\textbf{(b)}
For a given LLM, ReRound resolves midpoint ambiguity using a diffusion prior
(a U-Net) learned from $64{\times}64$ patches of all weight matrices in the
LLM.
Since this diffusion prior reflects the distribution of weights in the LLM,
ReRound uses it to guide the rounding.
In preparing for inference, the U-Net generates a collection of
$64{\times}64$ patches around $\widetilde{w}$.
The generated patches will be designated as the reconstructed 
$\widetilde{w}_{\mathrm{rec}}$.
Since $\widetilde{w}_{\mathrm{rec}}$ reflects the distribution of weights of
the LLM, ReRound applies RTN to $\widetilde{w}_{\mathrm{rec}}$, rather than
$\widetilde{w}$, when $\widetilde{w}_{\mathrm{rec}}$ is not far from
$\widetilde{w}$.
Thus, in this example, $\widetilde{w}=1.48$ under ReRound will be rounded to
the ceiling $2$ rather than the floor $1$, as shown in (a).
}
\label{fig:teaser}
\end{figure}

\section{Introduction}
\label{sec:intro}

Weight-only post-training quantization (PTQ) reduces the storage and
memory traffic of pretrained language models without retraining.
Accuracy under low-bit quantization has been improved through optimized
quantization parameters~\cite{FrantarAHA22,LinTTYXH25} and tensor
transformations~\cite{XiaoLSWDH23,AshkboosMCLCJAHH24}.
Once the scales, zero-points, group size, and quantized layers are fixed,
each quantized weight must still be assigned to its lower or upper
adjacent quantized integer.
We focus on this final rounding step.

Round-to-nearest (RTN) makes this assignment using scalar distance.
Its choice is clear when a quantized weight lies close to one adjacent
quantized integer.
Near the midpoint, however, rounding down and up incur nearly identical
error, and the decision depends on only a small distance difference.
Changing one such assignment incurs little additional scalar error.
Across a matrix, however, a small set of changes can alter the overall
rounding pattern applied by a linear layer.
Midpoint-ambiguous decisions therefore provide natural opportunities to
improve rounding at the matrix level.
Prior works show where RTN can be improved, but do not identify which
alternative rounding assignments should be chosen.

In this paper, we propose ReRound, a calibration-free framework that
revisits midpoint-ambiguous RTN decisions using information learned from
the pretrained LLM's own weights.
Quantization maps a range of full-precision values to the same low-bit
level, so the quantized integer alone does not reveal the original
full-precision value.
We train a conditional diffusion model on paired full-precision and
low-bit weight regions from the pretrained model to produce continuous
reconstructed weights from each low-bit observation.
The reconstructed weights are not deployed as model parameters.
Instead, they provide additional evidence for whether each original
weight should be rounded down or up.

The usefulness of this evidence depends on the margin of the original
RTN decision.
Near the midpoint, where the two assignments have similar scalar error,
the reconstructed weight can provide useful guidance.
Toward either adjacent quantized integer, RTN has a clearer preference
and should be preserved more often.
ReRound therefore uses a position-dependent tolerance metric determined
by the distance of each quantized weight from the midpoint.
This metric permits more reconstruction-guided changes within the
midpoint region and increasingly preserves RTN toward the adjacent
quantized integers.
A change is accepted only when the reconstructed weight favors the
assignment opposite to RTN and satisfies this tolerance metric.
Varying the tolerance parameter produces a small set of candidate
quantized integer weight matrices with different
reconstruction-guided assignments.

These candidates differ in their rounding patterns across the complete
weight matrix.
We therefore compare them as complete matrices.
The selected candidate is the one whose de-quantized weight matrix has
leading singular values that most closely match those of the original
full-precision matrix.
This weight-only spectral criterion favors candidates that preserve the
dominant structure of the original layer.
Both diffusion-based reconstruction and spectral selection require no
activation samples, calibration text, or downstream labels.

The final model differs from RTN only in selected floor-or-ceiling
assignments.
The scales, zero-points, group size, and quantized layers remain
unchanged.
The same procedure can therefore refine standard RTN or be applied to
quantization parameters produced by another PTQ method.
All diffusion-based reconstruction and spectral selection are performed
offline.
Because ReRound changes only the quantized integer assignments, it
leaves the low-bit representation and inference procedure unchanged.

Our contributions are:
\begin{itemize}
\item We introduce ReRound, which trains conditional diffusion on
the pretrained model's weight regions and uses the reconstructed
continuous weights to guide midpoint-ambiguous floor-or-ceiling
decisions without activation or text calibration data.

\item We propose a position-dependent tolerance metric that controls
where reconstruction-guided changes are accepted, together with a
weight-only spectral criterion that selects the final quantized
integer weight matrix.

\item ReRound consistently improves RTN at 3 and 4 bits, preserves the
same low-bit representation and inference procedure, and can be applied
to quantization parameters produced by another PTQ method.

\end{itemize}

\section{Related Work}
\label{sec:related_work}

\subsection{LLM Post-Training Quantization}
LLM PTQ improves low-bit accuracy by optimizing
quantization parameters or transforming weights and activations before
quantization.
GPTQ~\cite{FrantarAHA22} uses approximate second-order information to
reduce weight error, AWQ~\cite{LinTTYXH25} protects
activation-salient channels, and OmniQuant~\cite{ShaoCZXXLZPQL24}
learns clipping ranges and equivalent transformations from calibration
data.
SmoothQuant~\cite{XiaoLSWDH23} redistributes activation
outliers, whereas QuaRot~\cite{AshkboosMCLCJAHH24},
SpinQuant~\cite{LiuZFSCKCTB25}, and
FlatQuant~\cite{SunLBBZYYHJJ25} transform tensors into more
quantization-friendly representations.
Without calibration data, HQQ~\cite{BadriS23} optimizes a
weight-only objective, and SINQ~\cite{MullerBZCBC26} derives
dual-axis scales from matrix statistics.
These methods improve the quantization setup before the final rounding
step.

\subsection{Rounding Assignment Optimization}
With fixed quantization parameters, assigning each quantized weight to
its lower or upper adjacent quantized integer can still affect model
accuracy.
AdaRound~\cite{NagelABLB20} learns these assignments through local
output reconstruction, whereas FlexRound~\cite{LeeKKL23} introduces
element-wise divisive parameters for more flexible rounding.
SignRound~\cite{ChengZSCHLL24} jointly optimizes rounding and
clipping with signed gradients.
SignRoundV2~\cite{ChengZGS25} extends this approach with
gradient-informed layer sensitivity and quantization-scale
initialization for low-bit PTQ.
CafeQ~\cite{SunBKTSDT25} combines learned transformations with
calibration-free adaptive rounding under a weight-space proxy.
These methods optimize rounding through reconstruction or proxy
objectives, whereas ReRound uses reconstructed weights as a learned
prior for midpoint-ambiguous rounding decisions.

\subsection{Diffusion and Weight Priors}
Diffusion models have been used as priors for inverse problems by
incorporating observations into the reverse process.
DPS~\cite{ChungKMKY23} addresses noisy and nonlinear measurements,
and QCS-SGM~\cite{MengK22} and SIM-DMIS~\cite{TangCXL25} adapt
diffusion priors to quantized signal reconstruction.
For image dequantization, Vavilala et al.~\cite{VavilalaSF25}
reconstruct continuous colors from quantized representations.
Diff-OneBit~\cite{ChenL26} further combines a diffusion prior with data
consistency for reconstruction from 1-bit measurements.
In weight space, Neural Network Diffusion~\cite{WangTZYXZZDLY24} and
D2NWG~\cite{SoroALJCHH25} generate neural-network parameters.
Dravid et al.~\cite{DravidGWAWEA24} introduce weights2weights
for sampling, editing, and inverting customized diffusion models.
ReRound instead learns a diffusion prior from the pretrained LLM's own
weights and uses the reconstructed weights to guide midpoint-ambiguous
rounding decisions for low-bit weight-only PTQ.

\section{Method: ReRound}
\label{sec:method}

ReRound revisits the final rounding assignments under a fixed
quantization setup.
For each pretrained LLM, a conditional diffusion model is trained once
to produce continuous reconstructed weights
(Section~\ref{sec:weight_recovery}).
These reconstructed weights guide rounding decisions affected by
midpoint ambiguity and produce candidate quantized integer weight
matrices using a position-dependent tolerance metric
(Section~\ref{sec:recovery_guided_candidates}).
For each matrix, the candidate whose de-quantized weight matrix has
leading singular values that best match those of the full-precision
matrix is selected
(Section~\ref{sec:spectral_selection}).

\subsection{Midpoint-Ambiguous RTN Decisions}
\label{sec:rtn_margin}

Let $W\in\mathbb{R}^{d_{\mathrm{out}}\times d_{\mathrm{in}}}$ be a
full-precision weight matrix.
Consider one quantization group with scale $\Delta$ and zero-point $z$;
the same derivation applies independently to every group.
For a weight $w\in W$, define its quantized weight as
\begin{equation}
\tilde{w}
=
\frac{w}{\Delta}+z
=
\ell+r,
\qquad
\ell=\lfloor\tilde{w}\rfloor,
\quad
u=\ell+1,
\quad
r\in[0,1).
\label{eq:quantized_coordinate}
\end{equation}
RTN selects the closer adjacent quantized integer:
\begin{equation}
q_{\mathrm{rtn}}
=
\arg\min_{q\in\{\ell,u\}}
(\tilde{w}-q)^2.
\label{eq:rtn_assignment}
\end{equation}
The lower and upper assignments incur
\begin{equation}
\begin{aligned}
e_{\ell}(r)
&=
\left(w-\Delta(\ell-z)\right)^2
=
\Delta^2r^2,
\\
e_u(r)
&=
\left(w-\Delta(u-z)\right)^2
=
\Delta^2(1-r)^2.
\end{aligned}
\label{eq:adjacent_rounding_errors}
\end{equation}
Thus, flipping the RTN assignment incurs the additional cost
\begin{equation}
c_{\mathrm{flip}}(r)
=
\left|e_u(r)-e_{\ell}(r)\right|
=
\Delta^2|1-2r|.
\label{eq:rtn_flip_cost}
\end{equation}
The cost is smallest at the midpoint $r=0.5$ and increases toward
either adjacent quantized integer.
ReRound accordingly gives reconstructed weights greater influence near
the midpoint and increasingly preserves RTN away from it.

\begin{figure}[t]
\centering
\includegraphics[width=0.9\columnwidth]{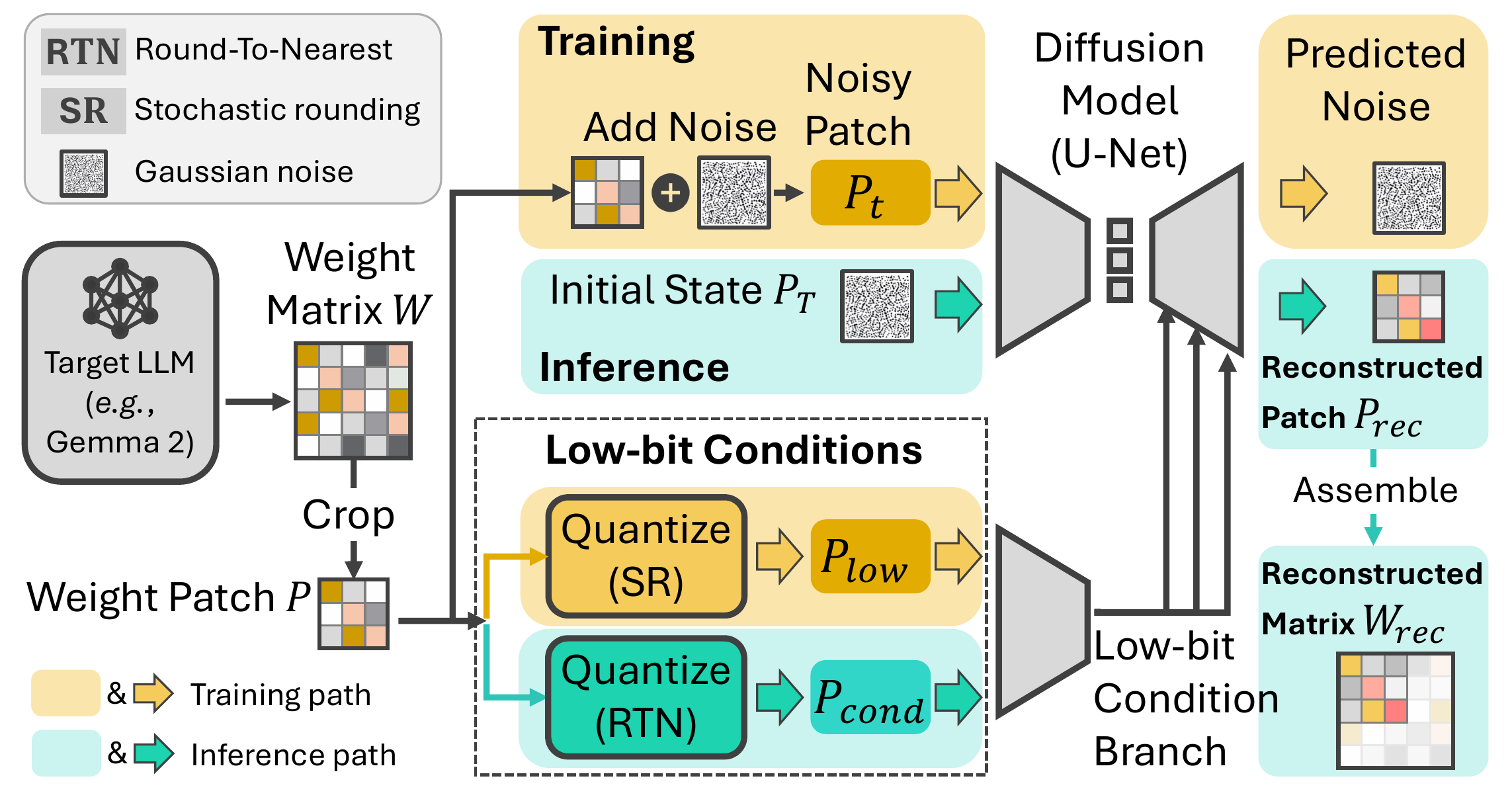}
\caption{
\textbf{Conditional weight reconstruction.}
\textcolor[HTML]{E2AC02}{During training}, the diffusion model predicts
noise on a full-precision patch $P$, conditioned on its stochastic
low-bit version $P_{\mathrm{low}}$.
\textcolor[HTML]{28ACA0}{During inference}, reverse diffusion is
conditioned on the deterministic RTN patch $P_{\mathrm{cond}}$.
The reconstructed patches form $W_{\mathrm{rec}}$, used only to guide
rounding.
}
\label{fig:diffusion}
\end{figure}

\subsection{Conditional Weight Reconstruction}
\label{sec:weight_recovery}

The flip cost in Eq.~\eqref{eq:rtn_flip_cost} identifies where an RTN
assignment can be changed at low scalar cost, but not whether the
alternative assignment is supported by the joint weight structure.
For each pretrained LLM, ReRound trains one conditional diffusion model
on local patches from its full-precision weight matrices.
The model jointly reconstructs each patch from its low-bit observation,
providing a prior learned from the pretrained LLM's own weights.

Let $P$ denote a patch cropped from a full-precision weight matrix $W$.
During training, its low-bit condition is formed by group-wise
stochastic quantization:
\begin{equation}
P_{\mathrm{low}}
=
Q_{\mathrm{sr}}(P;\Delta_c,z_c),
\label{eq:stochastic_condition}
\end{equation}
where $\Delta_c$ and $z_c$ are the corresponding scale and zero-point.
For an entry $p\in P$, define its quantized weight as
$\tilde{p}=p/\Delta_c+z_c=\ell+r$, where $r\in[0,1)$.
Then $Q_{\mathrm{sr}}$ selects the quantized integer $\ell$ with
probability $1-r$ and the quantized integer $\ell+1$ with probability
$r$.
Only the condition is quantized; the reconstruction target remains the
full-precision patch $P$.
ReRound models the conditional distribution
$p_{\theta}(P\mid P_{\mathrm{low}})$ using conditional
diffusion~\cite{VavilalaSF25}.

At diffusion step $t$, Gaussian noise
$\epsilon\sim\mathcal{N}(0,I)$ produces
\begin{equation}
P_t
=
\sqrt{\bar{\alpha}_t}\,P
+
\sqrt{1-\bar{\alpha}_t}\,\epsilon,
\label{eq:diffusion_forward}
\end{equation}
where $\bar{\alpha}_t$ determines the noise level.
The denoiser receives $P_t$, $t$, and $P_{\mathrm{low}}$ and predicts
the added noise:
\begin{equation}
\mathcal{L}_{\mathrm{diff}}
=
\mathbb{E}_{P,t,\epsilon}
\left[
\left\|
\epsilon-
\epsilon_{\theta}(P_t,t,P_{\mathrm{low}})
\right\|_2^2
\right].
\label{eq:diffusion_objective}
\end{equation}

During diffusion inference, let $\{P_k\}_{k=1}^{K}$ be the patches
cropped from a weight matrix $W$ of the pretrained LLM.
Each patch is conditioned on its deterministic RTN quantization:
\begin{equation}
P_{\mathrm{cond},k}
=
Q_{\mathrm{rtn}}(P_k;\Delta_c,z_c).
\label{eq:recovery_condition}
\end{equation}
Starting from $P_T\sim\mathcal{N}(0,I)$, the conditional reverse process
produces
\begin{equation}
P_{\mathrm{rec},k}
=
\mathcal{R}_{\theta}(P_{\mathrm{cond},k}),
\label{eq:recovered_weight_patch}
\end{equation}
where $\mathcal{R}_{\theta}$ denotes reverse diffusion from step $T$ to
step $0$.
The reconstructed patches are returned to their original locations to
form the reconstructed matrix $W_{\mathrm{rec}}$.
Because each patch is reconstructed jointly, each reconstructed entry
depends on its low-bit observation and the surrounding entries in the
patch.
The reconstructed matrix is used only to guide rounding and is not
deployed as the final weight matrix.

\subsection{Reconstruction-Guided Rounding}
\label{sec:recovery_guided_candidates}

\begin{figure}[ht!]
\centering
\includegraphics[width=0.9\columnwidth]{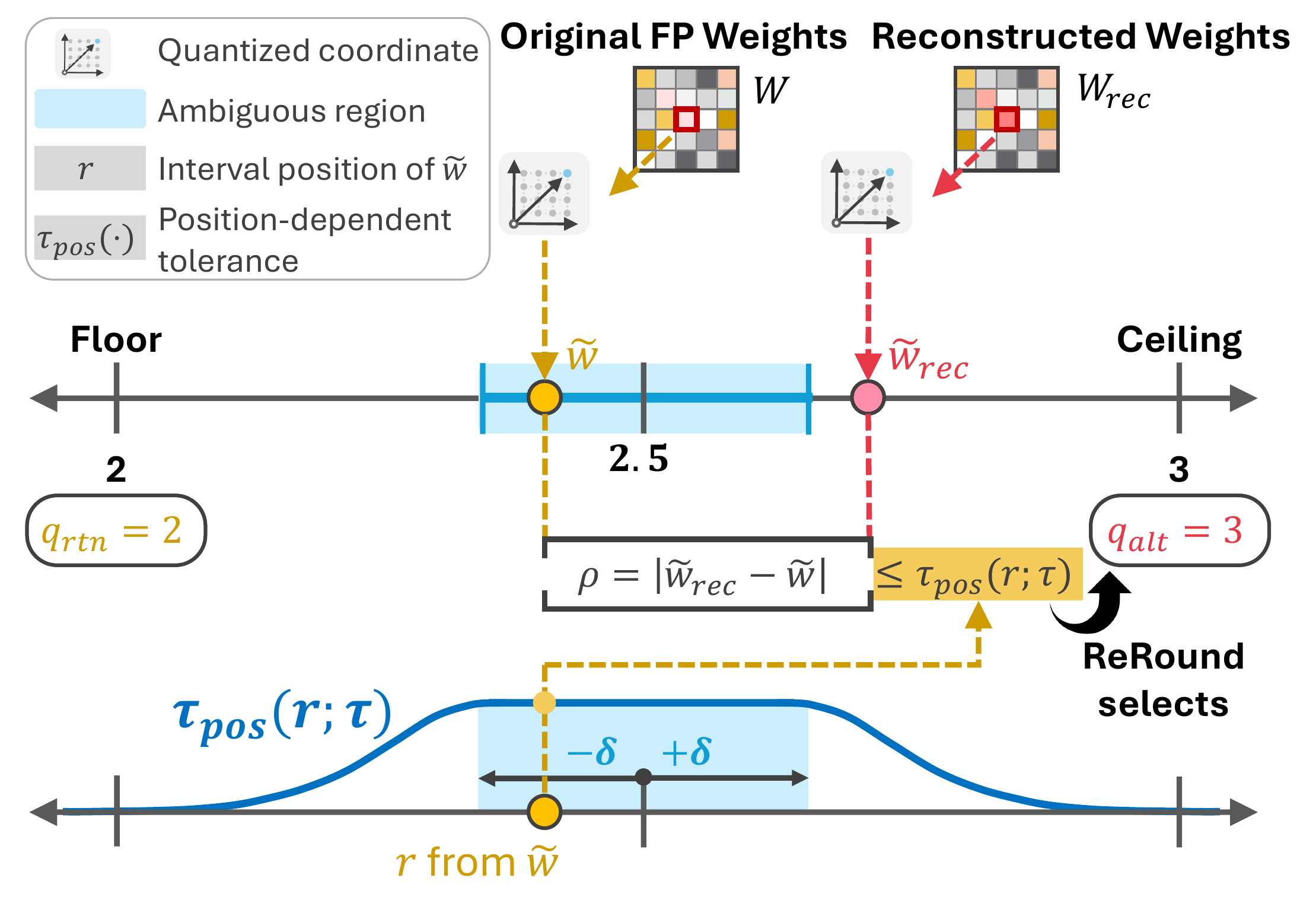}
\caption{
\textbf{Reconstruction-guided rounding.}
The original quantized weight $\tilde{w}$ determines the RTN assignment
$q_{\mathrm{rtn}}$ and the opposite adjacent assignment
$q_{\mathrm{alt}}$.
The reconstructed weight in quantized space,
$\tilde{w}_{\mathrm{rec}}$, proposes a change when
$q_{\mathrm{rec}}=q_{\mathrm{alt}}$.
The proposal is accepted when its deviation from $\tilde{w}$ lies
within the position-dependent tolerance metric
$\tau_{\mathrm{pos}}(r;\tau)$; otherwise, RTN is retained.
}
\label{fig:reround}
\end{figure}

ReRound uses the reconstructed weights to propose alternative
floor-or-ceiling assignments under the fixed PTQ quantization
parameters.
Let $w$ and $w_{\mathrm{rec}}$ be corresponding entries of $W$ and the
reconstructed matrix $W_{\mathrm{rec}}$.
Using the fixed scale $\Delta$ and zero-point $z$, we map both to the
same quantized weight space:
\begin{equation}
\tilde{w}
=
\frac{w}{\Delta}+z,
\qquad
\tilde{w}_{\mathrm{rec}}
=
\frac{w_{\mathrm{rec}}}{\Delta}+z.
\label{eq:recovered_quantized_coordinate}
\end{equation}
Using $\ell$, $u$, and $r$ from
Eq.~\eqref{eq:quantized_coordinate}, define
\begin{equation}
\begin{aligned}
q_{\mathrm{rtn}}
&=
\operatorname{round}(\tilde{w}),
&
q_{\mathrm{alt}}
&=
\ell+u-q_{\mathrm{rtn}},
\\
q_{\mathrm{rec}}
&=
\operatorname{round}(\tilde{w}_{\mathrm{rec}}).
\end{aligned}
\label{eq:recovery_assignments}
\end{equation}
The reconstructed weight proposes changing RTN only when
$q_{\mathrm{rec}}=q_{\mathrm{alt}}$.
Thus, reconstruction-guided rounding can only switch RTN to the
opposite adjacent quantized integer.
Whether the proposal is accepted depends on the original quantized
weight's position within its quantization interval.
Define its distance to the nearest quantized integer as
\begin{equation}
d(r)
=
\min\{r,1-r\}
\in[0,0.5].
\label{eq:quantized_position}
\end{equation}
Here, $d(r)=0.5$ at the midpoint and decreases toward zero near either
adjacent quantized integer.
By Eq.~\eqref{eq:rtn_flip_cost}, for a fixed scale $\Delta$, larger
$d(r)$ corresponds to a lower RTN flip cost.

The position-dependent tolerance metric for a reconstruction-guided
proposal depends on $d(r)$.
Given a tolerance parameter $\tau$, ReRound defines
\begin{equation}
\tau_{\mathrm{pos}}(r;\tau)
=
\tau
\exp
\left[
-\beta
\left(
\frac{
[0.5-\delta-d(r)]_{+}
}{
0.5-\delta
}
\right)^2
\right],
\label{eq:position_dependent_tolerance}
\end{equation}
where $[x]_{+}=\max\{x,0\}$.
The parameter $\delta$ defines the midpoint region
$|r-0.5|\leq\delta$, where
$\tau_{\mathrm{pos}}(r;\tau)=\tau$.
Outside this region, $\beta$ controls how quickly the tolerance metric
decreases toward either adjacent quantized integer.

The reconstruction deviation, measured in quantization-step units, is
\begin{equation}
\rho(w,w_{\mathrm{rec}})
=
\left|
\tilde{w}_{\mathrm{rec}}-\tilde{w}
\right|
=
\frac{|w_{\mathrm{rec}}-w|}{|\Delta|}.
\label{eq:recovery_coordinate_deviation}
\end{equation}
The candidate assignment for tolerance parameter $\tau$ is
\begin{equation}
q_{\tau}
=
\begin{cases}
q_{\mathrm{alt}},
&
q_{\mathrm{rec}}=q_{\mathrm{alt}}
\quad\text{and}\quad
\rho(w,w_{\mathrm{rec}})
\leq
\tau_{\mathrm{pos}}(r;\tau),
\\
q_{\mathrm{rtn}},
&
\text{otherwise}.
\end{cases}
\label{eq:reround_candidate_rule}
\end{equation}

Applying Eq.~\eqref{eq:reround_candidate_rule} element-wise produces
one candidate quantized integer weight matrix $Q_{\tau}$ for each
tolerance parameter $\tau\in\mathcal{T}$.
Different tolerance parameters produce different rounding candidates,
while all other quantization settings remain fixed.
The next section selects one candidate for each weight matrix.

\subsection{Candidate Selection via Spectral Preservation}
\label{sec:spectral_selection}

\begin{figure}[t]
\centering
\includegraphics[width=0.9\columnwidth]{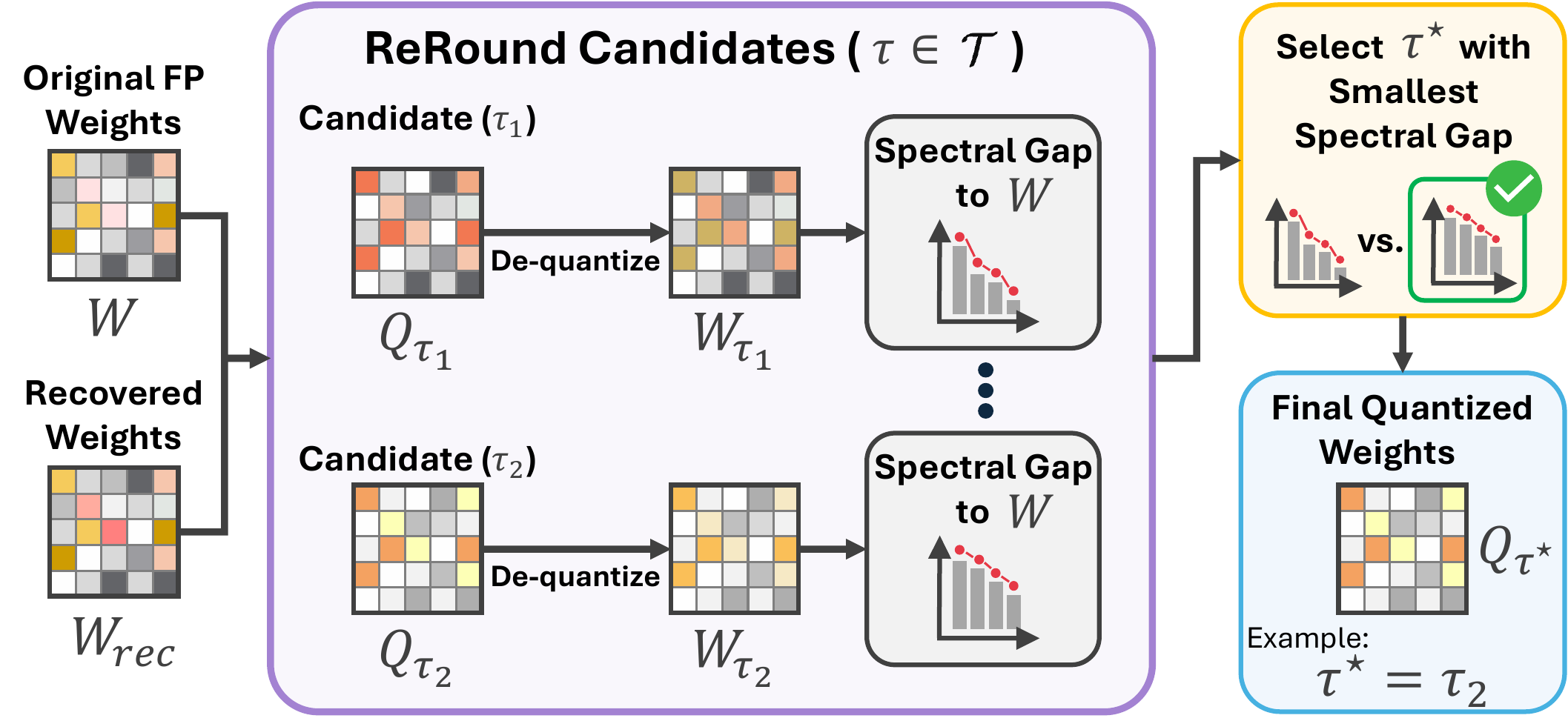}
\caption{
\textbf{Candidate selection via spectral preservation.}
Each candidate quantized integer weight matrix $Q_{\tau}$ is
de-quantized to obtain the candidate weight matrix $W_{\tau}$.
ReRound compares its leading singular values with those of the
full-precision matrix $W$ and selects the candidate with the smallest
spectral discrepancy.
The illustration selects $\tau^\star=\tau_2$.
}
\label{fig:reround_selection}
\end{figure}

\begin{table*}[!t]
\centering
\small
\setlength{\tabcolsep}{2.6pt}
\setlength{\TableWidth}{0.85\linewidth}

\resizebox{\TableWidth}{!}{
\begin{tabular}{@{}c@{\hspace{5pt}}l ccccA ccccA ccccA@{}}
\toprule
& & \multicolumn{5}{c}{\textbf{Gemma 2 2B}}
& \multicolumn{5}{c}{\textbf{Gemma 3 1B}}
& \multicolumn{5}{c}{\textbf{Qwen3 1.7B}} \\
\cmidrule(lr){3-7} \cmidrule(lr){8-12} \cmidrule(lr){13-17}
& Method
& WinoGrande$\uparrow$ & PIQA$\uparrow$ & BoolQ$\uparrow$ & SIQA$\uparrow$ & Avg.$\uparrow$
& WinoGrande$\uparrow$ & PIQA$\uparrow$ & BoolQ$\uparrow$ & SIQA$\uparrow$ & Avg.$\uparrow$
& WinoGrande$\uparrow$ & PIQA$\uparrow$ & BoolQ$\uparrow$ & SIQA$\uparrow$ & Avg.$\uparrow$ \\
\midrule

& 16-bit
& 69.0 & 79.3 & 72.7 & 51.4 & 68.1
& 58.8 & 74.9 & 66.5 & 42.9 & 60.8
& 60.9 & 72.6 & 77.5 & 45.2 & 64.1 \\

\midrule

\multirow{6}{*}{\bitslabel{W4A16}}
& RTN channel-wise
& 66.2 & 75.6 & 70.2 & 45.6 & 64.4
& 57.2 & 72.1 & 44.9 & 39.8 & 53.5
& 56.5 & 68.4 & 72.3 & 40.1 & 59.3 \\

& RTN group-wise, $G=128$
& 67.7 & 78.1 & 69.6 & 50.3 & 66.4
& 57.9 & 72.4 & 62.6 & 42.3 & 58.8
& 57.6 & \textbf{70.4} & 77.7 & 42.9 & 62.2 \\

& HQQ, $G=128$
& \textbf{68.8} & 78.2 & 71.4 & \textbf{50.8} & 67.3
& 58.9 & \textbf{73.9} & 54.7 & \textbf{42.9} & 57.6
& \textbf{59.0} & 69.0 & 75.0 & 42.4 & 61.3 \\

& BNB (FP4)
& 67.6 & 77.7 & 61.7 & 50.6 & 64.4
& 58.4 & 73.5 & 55.5 & 42.1 & 57.4
& 57.8 & 68.9 & 76.7 & 43.1 & 61.6 \\

& Hadamard + RTN, $G=128$
& 68.0 & 78.0 & 58.3 & 49.6 & 63.5
& 58.4 & 73.4 & \textbf{63.1} & 42.4 & \textbf{59.3}
& 57.1 & 66.5 & 72.4 & 41.3 & 59.3 \\

& \textbf{ReRound}
& 68.7 & \textbf{78.6} & \textbf{72.0} & 50.1 & \textbf{67.4}
& \textbf{59.3} & 73.3 & 61.9 & 42.6 & \textbf{59.3}
& 58.6 & 70.3 & \textbf{78.3} & \textbf{43.2} & \textbf{62.6} \\

\midrule

\multirow{5}{*}{\bitslabel{W3A16}}
& RTN channel-wise
& 51.3 & 55.4 & 61.5 & 34.3 & 50.6
& 50.2 & 52.6 & 47.3 & 33.9 & 46.0
& 50.5 & 53.1 & 40.1 & 34.4 & 44.5 \\

& RTN group-wise, $G=128$
& 63.2 & 73.5 & 68.3 & 42.6 & 61.9
& 55.5 & 65.9 & 60.9 & \textbf{38.9} & 55.3
& \textbf{53.4} & 62.5 & 62.9 & \textbf{37.5} & 54.1 \\

& HQQ, $G=128$
& 62.3 & 73.5 & 60.9 & 44.3 & 60.2
& 55.6 & \textbf{67.4} & 60.2 & 38.1 & 55.3
& 48.5 & 54.4 & 51.4 & 34.2 & 47.1 \\

& Hadamard + RTN, $G=128$
& 61.6 & \textbf{76.1} & 63.8 & \textbf{46.0} & 61.9
& 55.0 & \textbf{67.4} & 48.2 & 37.7 & 52.1
& 49.6 & 54.7 & 48.2 & 34.3 & 46.7 \\

& \textbf{ReRound}
& \textbf{63.8} & 73.1 & \textbf{68.6} & 42.6 & \textbf{62.0}
& \textbf{56.2} & 65.5 & \textbf{61.3} & \textbf{38.9} & \textbf{55.5}
& 52.5 & \textbf{62.7} & \textbf{64.6} & 36.9 & \textbf{54.2} \\

\bottomrule
\end{tabular}
}

\caption{
Calibration-free weight-only PTQ on Gemma 2 2B, Gemma 3 1B, and Qwen3 1.7B.
All group-wise methods use $G=128$, and the \texttt{lm\_head} is quantized for all methods.
ReRound shares the scales and zero-points of group-wise RTN and changes
only the rounding assignments.
Avg. reports the mean of the four task accuracies.
Best results for each model at each bit width are shown in bold.
}
\label{tab:sota_w4_w3_gemma_qwen}
\end{table*}

Reconstruction-guided rounding produces one candidate quantized integer
weight matrix $Q_{\tau}$ for each tolerance parameter
$\tau\in\mathcal{T}$.
ReRound selects the candidate whose de-quantized weight matrix has
leading singular values that most closely match those of the
full-precision matrix $W$.
These values summarize the strengths of the matrix's dominant
transformation modes, providing a matrix-level criterion without
requiring activation data.

Figure~\ref{fig:reround_selection} illustrates the selection process.
For each tolerance parameter $\tau\in\mathcal{T}$, the candidate
quantized integer weight matrix $Q_\tau$ is de-quantized using the fixed
PTQ parameters:
\begin{equation}
W_{\tau}
=
\operatorname{Dequant}
\left(
Q_{\tau};\Delta,z
\right),
\label{eq:candidate_dequantization}
\end{equation}
where $\Delta$ and $z$ denote the fixed scales and zero-points, applied
group-wise when needed.
The candidate is de-quantized only for spectral comparison.

Let $m=\min\{d_{\mathrm{out}},d_{\mathrm{in}}\}$ be the smaller
dimension of $W$.
We compare the leading $k$ singular values, where
$k=\min\{m,128,\max(32,\lfloor m/16\rfloor)\}$.
This choice adapts $k$ to the matrix size and uses at most $128$
singular values.
Let $\sigma_{1:k}(W)$ and $\sigma_{1:k}(W_{\tau})$ denote their
leading $k$ singular values, ordered from largest to smallest.
\begin{equation}
\mathcal{D}_{\mathrm{spec}}(\tau)
=
\frac{
\left\|
\sigma_{1:k}(W_{\tau})
-
\sigma_{1:k}(W)
\right\|_2
}{
\left\|
\sigma_{1:k}(W)
\right\|_2
+
\epsilon
},
\label{eq:spectral_discrepancy}
\end{equation}
where $\epsilon>0$ ensures numerical stability.
ReRound selects
\begin{equation}
\tau^{\star}
=
\arg\min_{\tau\in\mathcal{T}}
\mathcal{D}_{\mathrm{spec}}(\tau).
\label{eq:spectral_selection}
\end{equation}
The selected $Q_{\tau^\star}$ provides the final quantized integer
assignments and, together with the fixed scales and zero-points, defines
the quantized representation of $W$.

\section{Experiments}
\label{sec:experiments}

\subsection{Experimental Setup}
\label{sec:experimental_setup}

We summarize the main settings below; further implementation details
are provided in the supplementary material.

\paragraph{Models and evaluation.}
Our full baseline comparison includes Gemma 2
2B~\cite{TeamRPSHBHMSRR24}, Gemma 3 1B~\cite{TeamKFPVMPRM25},
Qwen3 1.7B~\cite{YangLAYZHBGYHL25}, OLMo 2
1B~\cite{TeamWSGLABGHJ25}, and SmolLM2
1.7B~\cite{AllalLBMPTMKLS25}.
We further evaluate whether ReRound consistently improves RTN across
Llama 3.2 1B~\cite{GrattafioriDJKD24}, Pythia
1.4B~\cite{BidermanSABHKPR23}, and Phi-2
2.7B~\cite{JavaheripiB23}.
We evaluate zero-shot accuracy on WinoGrande~\cite{SakaguchiBBC20},
PIQA~\cite{BiskZLGC20}, BoolQ~\cite{ClarkLCK0T19}, and
SIQA~\cite{SapRCBC19} using the LM Evaluation
Harness~\cite{eval-harness}.
For SINQ~\cite{MullerBZCBC26}, we adopt its quantization scales and
also report perplexity on WikiText-2~\cite{MerityX0S17} and
C4~\cite{RaffelSRLNMZLL20}.

\begin{table*}[!t]
\centering
\small
\setlength{\tabcolsep}{4.0pt}
\setlength{\TableWidth}{0.68\linewidth}

\resizebox{\TableWidth}{!}{
\begin{tabular}{@{}c@{\hspace{5pt}}l ccccA ccccA@{}}
\toprule
& & \multicolumn{5}{c}{\textbf{OLMo 2 1B}}
& \multicolumn{5}{c}{\textbf{SmolLM2 1.7B}} \\
\cmidrule(lr){3-7} \cmidrule(lr){8-12}
& Method
& WinoGrande$\uparrow$ & PIQA$\uparrow$ & BoolQ$\uparrow$ & SIQA$\uparrow$ & Avg.$\uparrow$
& WinoGrande$\uparrow$ & PIQA$\uparrow$ & BoolQ$\uparrow$ & SIQA$\uparrow$ & Avg.$\uparrow$ \\
\midrule

& 16-bit
& 64.6 & 75.2 & 62.3 & 43.4 & 61.4
& 66.1 & 77.3 & 72.5 & 44.3 & 65.1 \\

\midrule

\multirow{6}{*}{\bitslabel{W4A16}}
& RTN channel-wise
& 59.4 & 72.4 & 63.2 & 42.2 & 59.3
& 58.1 & 72.2 & 64.3 & 40.1 & 58.7 \\

& RTN group-wise, $G=128$
& 63.5 & 74.9 & 51.7 & 42.3 & 58.1
& 63.5 & 76.2 & \textbf{68.3} & 42.8 & 62.7 \\

& HQQ, $G=128$
& \textbf{64.3} & 74.8 & 59.9 & 42.5 & 60.4
& 61.8 & 75.4 & 67.3 & 43.0 & 61.9 \\

& BNB (FP4)
& 61.9 & 74.7 & \textbf{64.3} & 43.4 & \textbf{61.1}
& 62.7 & 76.3 & 54.4 & \textbf{44.0} & 59.3 \\

& Hadamard + RTN, $G=128$
& 63.3 & \textbf{75.5} & 52.3 & 43.3 & 58.6
& 62.8 & 74.8 & \textbf{68.3} & 42.0 & 62.0 \\

& \textbf{ReRound}
& 63.9 & 75.4 & 54.4 & \textbf{43.7} & 59.4
& \textbf{65.1} & \textbf{76.7} & 67.3 & 42.3 & \textbf{62.9} \\

\midrule

\multirow{5}{*}{\bitslabel{W3A16}}
& RTN channel-wise
& 51.1 & 57.6 & 38.8 & 33.8 & 45.4
& 48.2 & 51.0 & 49.4 & 33.3 & 45.5 \\

& RTN group-wise, $G=128$
& \textbf{58.7} & 71.2 & 47.7 & 39.5 & 54.3
& \textbf{57.9} & 70.0 & 58.0 & 38.2 & 56.0 \\

& HQQ, $G=128$
& 56.3 & 67.1 & \textbf{58.0} & 36.4 & 54.4
& 53.4 & 60.8 & 58.9 & 37.8 & 52.8 \\

& Hadamard + RTN, $G=128$
& 55.6 & 68.4 & 42.1 & 35.5 & 50.4
& 51.5 & 59.7 & 45.0 & 36.1 & 48.1 \\

& \textbf{ReRound}
& 57.7 & \textbf{71.4} & 51.4 & \textbf{39.9} & \textbf{55.1}
& 57.5 & \textbf{71.0} & \textbf{60.6} & \textbf{38.3} & \textbf{56.9} \\

\bottomrule
\end{tabular}
}

\caption{
Calibration-free weight-only PTQ on OLMo 2 1B and SmolLM2 1.7B.
All group-wise methods use $G=128$, and all methods quantize
\texttt{lm\_head}.
ReRound shares the scales and zero-points of group-wise RTN and changes
only the rounding assignments.
Avg. is the mean across four tasks.
Best results for each model at each bit width are shown in bold.
}
\label{tab:sota_w4_w3_olmo_smol}
\end{table*}

\paragraph{Quantization settings.}
In the main RTN and ReRound experiments, we use asymmetric uniform
weight quantization.
W4A16 and W3A16 denote 4-bit and 3-bit weight quantization,
respectively, both with 16-bit activations.
All group-wise methods use group size $G=128$.
For all evaluated quantization methods, we quantize the transformer
linear layers and the output logit layer (\texttt{lm\_head}).
ReRound uses the same scales, zero-points, group size, and layer coverage
as group-wise RTN; only selected quantized integer assignments differ.
It therefore preserves the same low-bit representation and inference
procedure as RTN.

\paragraph{Baselines.}
The calibration-free baselines are channel-wise and group-wise RTN,
HQQ, BNB FP4~\cite{DettmersPHZ23}, Hadamard-transformed
RTN~\cite{AshkboosMCLCJAHH24,MullerBZCBC26}, and CafeQ.
BNB FP4 is evaluated only at 4 bits.
The 4-bit calibration-based baselines are GPTQ, AdaRound, and
SignRound.
GPTQ and AdaRound use 128 calibration samples from C4.

\paragraph{Diffusion training and inference.}
For each pretrained LLM, we train one conditional diffusion
model~\cite{VavilalaSF25} on two GPUs using full-precision patches
paired with 2-bit conditions generated by stochastic rounding (SR).
Diffusion inference uses one GPU to reconstruct $W_{\mathrm{rec}}$ from
deterministic 2-bit RTN conditions.
The same reconstructed matrix $W_{\mathrm{rec}}$ is reused for the
3- and 4-bit ReRound runs.
For SINQ, we instead train separate models for 3-bit and 4-bit
conditions, using SR during training and RTN during diffusion inference.

\paragraph{ReRound PTQ.}
For each target bit width, ReRound uses one GPU to map $W$ and
$W_{\mathrm{rec}}$ to the same quantized weight space, construct
candidate quantized integer weight matrices over
$\tau\in\mathcal{T}$, and select one candidate for each weight matrix
by matching the leading singular values of its de-quantized weight
matrix to those of $W$.
It changes at most $1\%$ of the RTN quantized integer assignments in
each matrix and uses the same formulation across models.
All diffusion-based reconstruction and ReRound PTQ stages use only the
pretrained model's weights, without activation or text calibration
samples.

\begin{table}[t]
\centering
\small
\setlength{\tabcolsep}{3.2pt}
\setlength{\TableWidth}{0.8\columnwidth}

\resizebox{\TableWidth}{!}{
\begin{tabular}{@{}l l ccccA@{}}
\toprule
Model & Method
& WinoGrande$\uparrow$ & PIQA$\uparrow$ & BoolQ$\uparrow$ & SIQA$\uparrow$ & Avg.$\uparrow$ \\
\midrule

\multirow{6}{*}{Gemma 2 2B}
& 16 bits
& 69.0 & 79.3 & 72.7 & 51.4 & 68.1 \\
\cmidrule(lr){2-7}
& GPTQ, $G=128$
& 66.5 & \textbf{78.6} & 70.9 & 50.8 & 66.7 \\
& AdaRound, $G=128$
& 68.2 & 78.5 & 70.6 & 50.2 & 66.9 \\
& SignRound, $G=128$
& 67.9 & 78.4 & 70.8 & 51.1 & 67.0 \\

& CafeQ
& 66.9 & 78.0 & 67.1 & \textbf{51.4} & 65.9 \\
& \textbf{ReRound}
& \textbf{68.7} & \textbf{78.6} & \textbf{72.0} & 50.1 & \textbf{67.4} \\

\midrule

\multirow{5}{*}{Gemma 3 1B}
& 16 bits
& 58.8 & 74.9 & 66.5 & 42.9 & 60.8 \\
\cmidrule(lr){2-7}
& GPTQ, $G=128$
& 58.0 & \textbf{73.9} & 56.8 & 42.3 & 57.8 \\
& AdaRound, $G=128$
& 59.1 & 73.0 & 61.4 & 42.0 & 58.9 \\
& SignRound, $G=128$
& 58.5 & 73.2 & \textbf{63.5} & 41.3 & 59.1 \\

& \textbf{ReRound}
& \textbf{59.3} & 73.3 & 61.9 & \textbf{42.6} & \textbf{59.3} \\

\bottomrule
\end{tabular}
}
\caption{
Comparison with calibration-based PTQ under W4A16 quantization on
Gemma 2 2B and Gemma 3 1B, with CafeQ additionally reported on
Gemma 2 2B.
GPTQ, AdaRound, and SignRound use calibration data, whereas CafeQ and
ReRound are calibration-free.
All group-wise methods use $G=128$, and all methods quantize
\texttt{lm\_head}.
Avg. is the mean accuracy across the four tasks.
Best quantized results for each model are shown in bold.
}
\label{tab:w4_calibrated_calibrationfree_gemma}
\end{table}

\begin{table}[t]
\centering
\scriptsize
\setlength{\tabcolsep}{2.2pt}
\resizebox{0.8\columnwidth}{!}{
\begin{tabular}{@{}c@{\hspace{5pt}}l ccc ccccA@{}}
\toprule
Bits & Method & Rec. & Pos. & Spec. & Wino. & PIQA & BoolQ & SIQA & Avg. \\
\midrule
\multirow{5}{*}{\bitslabel{W4A16}}
& RTN (group-wise, $G=128$) & -- & -- & -- & 63.5 & 74.9 & 51.7 & 42.3 & 58.1 \\
& Matched random flips    & --         & \checkmark & \checkmark & 63.5          & 75.3          & 51.8          & 43.3          & 58.5 \\
& Midpoint window   & \checkmark & --         & \checkmark & 63.8          & \textbf{75.6} & 50.7          & 43.6          & 58.4 \\
& Weight-MSE selection & \checkmark & \checkmark & --         & \textbf{64.6} & 75.5          & 50.6          & \textbf{44.0} & 58.7 \\
& ReRound              & \checkmark & \checkmark & \checkmark & 63.9          & 75.4          & \textbf{54.4} & 43.7          & \textbf{59.4} \\
\midrule
\multirow{5}{*}{\bitslabel{W3A16}}
& RTN (group-wise, $G=128$) & -- & -- & -- & 58.7 & 71.2 & 47.7 & 39.5 & 54.3 \\
& Matched random flips  & --         & \checkmark & \checkmark & 57.4          & 70.2          & 49.3          & 39.3          & 54.1 \\
& Midpoint window   & \checkmark & --         & \checkmark & \textbf{59.0} & 71.2          & 48.0          & 39.6          & 54.5 \\
& Weight-MSE selection & \checkmark & \checkmark & --         & 58.7          & 71.2          & 47.7          & 39.5          & 54.3 \\
& ReRound              & \checkmark & \checkmark & \checkmark & 57.7          & \textbf{71.4} & \textbf{51.4} & \textbf{39.9} & \textbf{55.1} \\
\bottomrule
\end{tabular}}
\caption{\textbf{Component ablation on OLMo 2 1B.}
Each variant replaces one ReRound component while keeping the other two.
Rec., Pos., and Spec. denote reconstruction guidance,
position-dependent tolerance, and spectral selection.
Matched random flips make the same number of assignment changes as
ReRound at random and are averaged over ten seeds.
Midpoint window replaces position-dependent tolerance with
$[0.5-\delta,\,0.5+\delta]$, while weight-MSE selection replaces
spectral discrepancy with element-wise weight MSE.
All variants use the same group-wise W3/W4 quantization parameters,
$G=128$, 16-bit activations, and layer coverage.
Best results at each bit width are bolded.
}

\label{tab:component_ablation}
\end{table}

\subsection{Comparison with Calibration-Free PTQ}
\label{sec:calibration_free_results}

Tables~\ref{tab:sota_w4_w3_gemma_qwen}
and~\ref{tab:sota_w4_w3_olmo_smol} compare ReRound with
calibration-free weight-only PTQ methods across five LLMs.
ReRound improves its matched group-wise RTN baseline for every model at
both bit widths, with gains of $0.1$--$0.9$ points at 3 bits and
$0.2$--$1.3$ points at 4 bits.
Because the scales, zero-points, group size, and layer coverage are
fixed, these gains come solely from revising selected quantized integer
assignments.

At 3 bits, ReRound achieves the highest four-task average on every
model, showing the benefit of revisiting midpoint-ambiguous decisions
under aggressive quantization.
At 4 bits, it achieves or matches the best average on four of the five
models.
On OLMo 2 1B, HQQ and BNB FP4 obtain averages of $60.4$ and $61.1$, respectively, compared with $59.4$ for ReRound. Nevertheless, ReRound raises its matched group-wise RTN baseline from $58.1$ to $59.4$, the largest 4-bit gain in the comparison.
Thus, even when another quantization parameterization performs better, revising selected quantized integer assignments substantially improves the fixed RTN quantization setup.

\begin{table}[!t]
\centering
\small
\setlength{\tabcolsep}{4.0pt}
\setlength{\TableWidth}{0.99\columnwidth}

\resizebox{\TableWidth}{!}{
\begin{tabular}{@{}c@{\hspace{5pt}}l ccccA ccA@{}}
\toprule
& Method
& WinoGrande$\uparrow$ & PIQA$\uparrow$ & BoolQ$\uparrow$ & SIQA$\uparrow$ & Acc Avg.$\uparrow$
& Wiki2$\downarrow$ & C4$\downarrow$ & PPL Avg.$\downarrow$ \\
\midrule

& 16-bit
& 60.9 & 72.6 & 77.5 & 45.2 & 64.1
& 16.72 & 19.25 & 17.99 \\

\midrule

\multirow{3}{*}{\bitslabel{W4A16}}
& SINQ-scale RTN, $G=64$
& -- & -- & -- & -- & --
& 17.14 & 19.83 & 18.49 \\

& SINQ-scale RTN, $G=64$
& 59.0 & 71.5 & \textbf{75.1} & 42.9 & 62.1
& 17.19 & 19.87 & 18.53 \\

& \textbf{SINQ-scale ReRound}, $G=64$
& \textbf{59.7} & \textbf{71.7} & 74.8 & \textbf{44.0} & \textbf{62.6}
& \textbf{17.09} & \textbf{19.83} & \textbf{18.46} \\

\midrule

\multirow{3}{*}{\bitslabel{W3A16}}
& SINQ-scale RTN, $G=64$
& -- & -- & -- & -- & --
& 22.39 & 24.88 & 23.64 \\

& SINQ-scale RTN, $G=64$
& 54.1 & \textbf{67.4} & \textbf{66.1} & 40.6 & 57.1
& 23.27 & 25.25 & 24.26 \\

& \textbf{SINQ-scale ReRound}, $G=64$
& \textbf{55.6} & 66.8 & 65.9 & \textbf{40.9} & \textbf{57.3}
& \textbf{23.14} & \textbf{24.97} & \textbf{24.06} \\

\bottomrule
\end{tabular}
}
\caption{
Rounding with SINQ quantization scales on Qwen3 1.7B.
All quantized rows use group size $G=64$ and follow the layer coverage of SINQ, with the \texttt{lm\_head} left unquantized.
The SINQ-reported RTN perplexities are included for reference.
For the reproduced results, ReRound differs from RTN only in the rounding rule.
Acc. Avg. and PPL Avg. report the means over the four accuracy tasks and two perplexity datasets, respectively.
Bold compares the reproduced RTN and ReRound results.
}
\label{tab:sinq_scale_reround_qwen3_1p7b}
\end{table}

\subsection{Comparison with Calibration-Based PTQ}
\label{sec:calibrated_comparison}

Table~\ref{tab:w4_calibrated_calibrationfree_gemma} compares ReRound
with calibration-based PTQ methods under W4A16 quantization, together
with CafeQ on Gemma 2 2B.
ReRound achieves the highest four-task average among the quantized
methods on both models.
The strongest calibration-based method, SignRound, reaches averages of
$67.0$ on Gemma 2 2B and $59.1$ on Gemma 3 1B, whereas ReRound achieves
$67.4$ and $59.3$, respectively.
Thus, ReRound surpasses the best calibration-based result by $0.4$ and
$0.2$ points without using activation or text calibration data.
These results show that revisiting midpoint-ambiguous rounding
decisions with a weight prior learned from the pretrained model's own
weights can compete effectively with calibration-data-driven PTQ.

\subsection{Ablation of ReRound Components}
\label{sec:component_ablation}

Table~\ref{tab:component_ablation} isolates reconstruction guidance,
the position-dependent tolerance metric, and spectral selection on
OLMo 2 1B.
Each ablation replaces one component while retaining the other two.
Matched random flips remove reconstruction guidance by randomly
choosing the same number of quantized integer assignment changes as
ReRound.
Midpoint window replaces the position-dependent tolerance metric with
$[0.5-\delta,,0.5+\delta]$, while weight-MSE selection uses
element-wise weight MSE instead of spectral discrepancy.
Full ReRound achieves the highest four-task average at both bit widths,
improving group-wise RTN by $1.3$ points at W4 and $0.8$ points at W3.
Replacing any component lowers the average accuracy.

\subsection{Alternative Quantization Parameters}
\label{sec:sinq_scales}

Table~\ref{tab:sinq_scale_reround_qwen3_1p7b} evaluates ReRound using
quantization parameters generated by another PTQ method. We use SINQ
parameters and compare RTN with ReRound under the same reproduced
setting, including SINQ's reported RTN perplexities for reference.
At 4 bits, ReRound increases the four-task average from $62.1$ to
$62.6$ and reduces average perplexity from $18.53$ to $18.46$.
At 3 bits, it increases the average from $57.1$ to $57.3$ and reduces
average perplexity from $24.26$ to $24.06$. Thus, ReRound also improves quantization setups from other PTQ methods.

\subsection{RTN Improvements on Additional Models}
\label{sec:rtn_improvements_across_models}

\begin{figure}[t]
\centering
\includegraphics[width=0.9\linewidth]{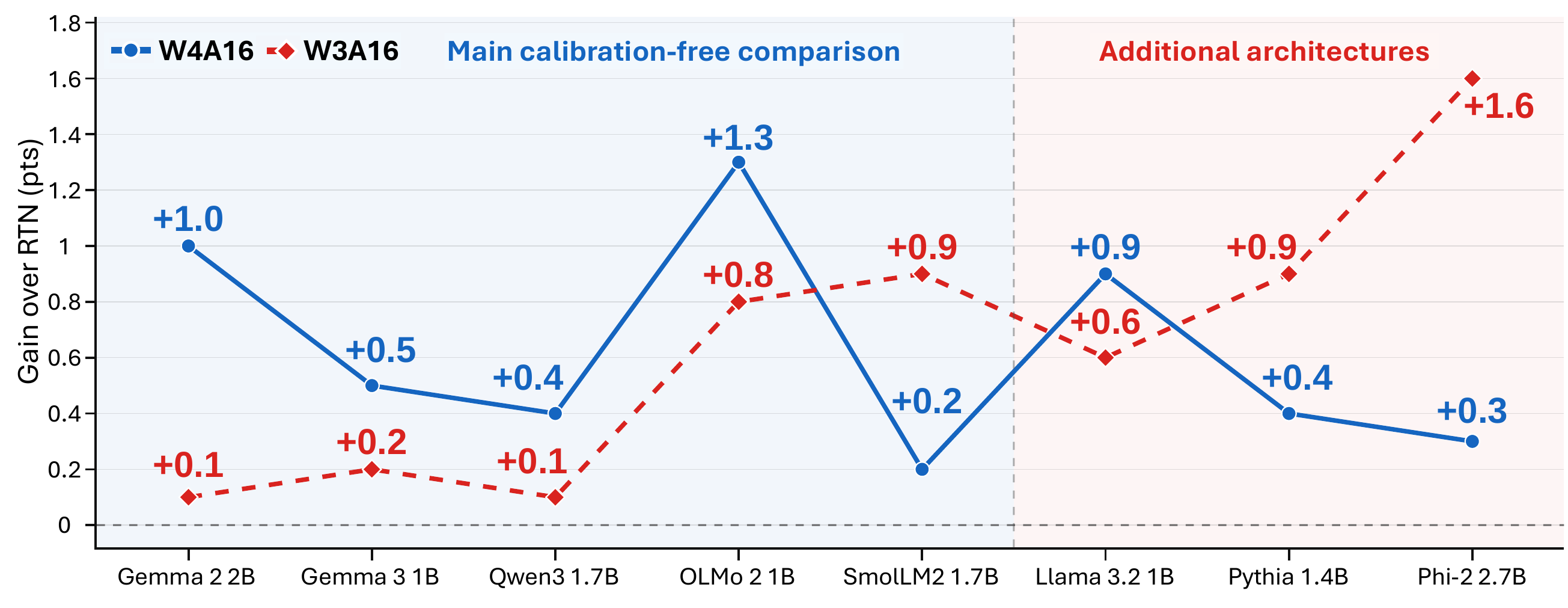}
\caption{
Average-accuracy gains of ReRound over group-wise RTN across eight
LLMs, averaged over WinoGrande, PIQA, BoolQ, and SIQA.
Both methods use identical quantization settings; positive values favor
ReRound.
}
\label{fig:rtn_gain_across_models}
\end{figure}

Figure~\ref{fig:rtn_gain_across_models} extends the five-model
comparison with Llama 3.2 1B, Pythia 1.4B, and Phi-2 2.7B.
ReRound improves group-wise RTN for every model at both 3 and 4 bits.
The gains range from $0.2$ to $1.3$ points at 4 bits and from $0.1$ to
$1.6$ points at 3 bits.
Because ReRound and RTN use the same scales, zero-points, group size,
and layer coverage, these differences arise solely from revising
selected quantized integer assignments.
The positive gains on the three additional models extend the same trend
beyond those used in the full calibration-free comparison.
Complete per-task results for these models are provided in the
supplementary material.

\begin{table}[t]
\centering
\small
\setlength{\tabcolsep}{4.5pt}
\resizebox{0.9\columnwidth}{!}{
\begin{tabular}{@{}lccc@{}}
\toprule
Model
& Diffusion Training (h)
& Diffusion Inference (h)
& ReRound PTQ (s)
\\
\midrule
Gemma 2 2B
& 2.52
& 9.55
& 105
\\
Gemma 3 1B
& 2.69
& 3.65
& 42
\\
Qwen3 1.7B
& 1.76
& 6.29
& 74
\\
OLMo 2 1B
& 1.77
& 9.32
& 57
\\
SmolLM2 1.7B
& 3.43
& 6.25
& 124
\\
\bottomrule
\end{tabular}
}

\caption{\textbf{Offline runtime of ReRound.}
Diffusion training and inference are performed once per model, and the
reconstructed matrix $W_{\mathrm{rec}}$ is reused for W3 and W4.
ReRound PTQ reports the mean runtime over the two target bit widths.
}
\label{tab:offline_cost}
\end{table}

\subsection{Offline Runtime}
\label{sec:offline_runtime}

Table~\ref{tab:offline_cost} reports the offline cost of ReRound.
Diffusion training is performed once per model on two GPUs, and
one-GPU diffusion inference produces the reconstructed matrix
$W_{\mathrm{rec}}$ for reuse across bit widths.
With this shared reconstruction, ReRound PTQ averages $42$ to $124$
seconds across W3 and W4.
All additional computation occurs before deployment; the resulting
quantized model uses the same low-bit representation and inference
procedure as RTN.

\section{Discussion and Limitations}
\label{sec:limitations}

ReRound revises RTN assignments while keeping scales and zero-points
fixed, so it can refine quantization parameters produced by other PTQ
methods. However, its candidate set remains constrained by these
parameters, motivating joint optimization of quantization parameters
and reconstruction-guided rounding. Although the approach may extend
beyond LLMs, we evaluate only small LLMs.

ReRound also requires a separate diffusion model for each pretrained
LLM, introducing model-specific offline cost. Reusing reconstructed
weights across bit widths reduces this cost, but scaling to many models
remains expensive. In addition, spectral selection operates only in
weight space and may not identify the best candidate for every
downstream task, while patch-based reconstruction may miss long-range
and cross-layer dependencies. Reducing reconstruction cost, transferring
weight priors across models, and using broader structural criteria are
promising directions.

\section{Conclusion}
\label{sec:conclusion}

We presented ReRound, a calibration-free framework that revisits
midpoint-ambiguous RTN decisions after quantization parameters are
fixed. ReRound learns a diffusion prior from the model's own weights,
uses it to guide floor-or-ceiling changes, and selects candidate
matrices through spectral preservation. Across diverse LLMs, it
consistently improves 3- and 4-bit weight-only PTQ without activation
or text calibration data. Because ReRound changes only selected
quantized integer assignments, the quantization parameters, low-bit
representation, and inference procedure remain unchanged. These
results show that structural patterns in pretrained weights can serve
as an effective prior for resolving ambiguous rounding decisions.

\appendix

\section{Quantization Notation}
\label{sec:supp_quantization_notation}

Let $w$ be a full-precision weight in a quantization group with scale
$\Delta$ and zero-point $z$.
Applying these fixed quantization parameters gives
\begin{equation}
\tilde{w}
=
\frac{w}{\Delta}+z
=
\ell+r,
\qquad
\ell=\lfloor\tilde{w}\rfloor,
\quad
u=\ell+1,
\quad
r\in[0,1).
\label{eq:quantized_coordinate}
\end{equation}
We refer to the continuous floating-point value $\tilde{w}$ before
rounding as the \emph{quantized floating-point weight}, or simply the
\emph{quantized weight}.
Rounding $\tilde{w}$ produces a discrete \emph{quantized integer}
$q\in\{\ell,u\}$.
Mapping $q$ back to the original weight space gives the corresponding
dequantized weight, $\Delta(q-z)$.

\section{Software Environment}
\label{sec:supp_software_environment}

We use the same software environment for diffusion-based weight
recovery, post-training quantization (PTQ), and language-model
evaluation.
PyTorch is compiled with CUDA 12.4.
Table~\ref{tab:software_versions} lists the main packages used in our
experiments.
For weight recovery, we follow the DeepFloyd IF Stage-II~\cite{DeepFloydTeam23, SahariaCSLWDGLKS22} implementation
used by Vavilala et al.~\cite{VavilalaSF25} and initialize the model
from the \texttt{IF-II-M-v1.0} pretrained checkpoint.

\begin{table}[t]
\centering
\small
\begin{tabular}{ll}
\hline
Package & Version \\
\hline
PyTorch & 2.5.1+cu124 \\
TorchVision & 0.20.1+cu124 \\
CUDA runtime & 12.4.127 \\
cuDNN & 9.1.0.70 \\
Accelerate & 1.1.1 \\
Transformers & 4.57.1 \\
Diffusers & 0.35.2 \\
Datasets & 5.0.0 \\
lm-eval & 0.4.11 \\
Hugging Face Hub & 0.36.0 \\
Tokenizers & 0.22.1 \\
\hline
\end{tabular}
\caption{Software packages and versions used for diffusion
weight recovery, PTQ, and language-model evaluation.}
\label{tab:software_versions}
\end{table}

\section{Diffusion Training and Inference Details}
\label{sec:supp_diffusion_details}

\paragraph{Diffusion architecture and conditioning.}
Following Vavilala et al.~\cite{VavilalaSF25}, we build the weight
recovery model from DeepFloyd IF Stage II
(\texttt{IF-II-M-v1.0}).
We add a ControlNet to the pretrained U-Net and use the low-bit weight
patch as the ControlNet condition.
During training, the U-Net remains frozen and only the ControlNet is
updated.
The ControlNet text-encoder pooling and projection layers are also kept
frozen.

Because weight recovery does not require natural-language information,
we use a precomputed T5 embedding of an empty prompt for all training
and inference runs.
Each weight patch and its low-bit condition contain one channel.
We repeat each of them across three channels to match the input format
of the pretrained diffusion model.

\paragraph{Weight-patch construction.}
For each target LLM, we extract its two-dimensional weight matrices,
including the language-model head.
We exclude token embeddings and normalization parameters.
The extracted weights are converted to FP32 and divided into
$64\times64$ patches.

Before they are given to the diffusion model, the weights in each
quantization group are normalized to the range $[0,1]$ using the
minimum and range of that group.
We use asymmetric group-wise quantization with a group size of 128 to
generate the low-bit conditions.

During training, we first randomly select a weight matrix and then
randomly crop a $64\times64$ patch from that matrix.
This allows the model to observe patches from different layers and
different locations throughout training.
If a matrix is smaller than the patch size, the patch is padded to
$64\times64$, and a validity mask identifies the original, non-padded
entries.
We also horizontally flip each training patch with probability $0.5$.

Each normalized full-precision target patch is paired with a 2-bit
condition generated by stochastic rounding (SR).
SR rounds each quantized weight to its lower or upper neighboring
integer according to its fractional part.
The resulting 2-bit patch is used only as the condition, while the
training target remains the normalized full-precision patch.

\paragraph{Optimization and hardware.}
We train a separate diffusion model for each target LLM using two
NVIDIA GeForce RTX 4090 GPUs.
Each model is trained for five epochs, with $1{,}000{,}000$ randomly
sampled patches per epoch.
The batch size is 32 per GPU, giving an effective batch size of 64
across the two GPUs, without gradient accumulation.

We optimize the ControlNet using AdamW with a constant learning rate of
$2\times10^{-5}$ and no learning-rate scheduler.
We use the default PyTorch AdamW settings:
$\beta_1=0.9$, $\beta_2=0.999$, $\epsilon=10^{-8}$, and weight decay
$0.01$.
The script-level random seed is set to 1.

\paragraph{Diffusion inference and matrix assembly.}
Diffusion inference is performed on one NVIDIA GeForce RTX 4090 GPU
with a batch size of 128.
Unlike training, inference uses deterministic 2-bit RTN conditions.
We use DDPM sampling with the \texttt{super27} timestep-respacing
schedule, which contains 27 reverse-diffusion steps.

During inference, we process all $64\times64$ patches of each weight
matrix in a fixed order.
We begin at the top-left corner, move from left to right, and then
continue to the next rows.
If the matrix size is not divisible by 64, we add a final patch aligned
with the bottom or right edge so that every weight is covered.

The diffusion model predicts each patch in the normalized $[0,1]$
space.
We then map the predicted values back to the original full-precision
weight space using the minimum and range of their corresponding
quantization groups.
Each recovered patch is written back to its original location.
After all patches are processed, they are combined to form the complete
recovered weight matrix $W_{\mathrm{rec}}$.

For ReRound, we train one diffusion model per target LLM using 2-bit SR
conditions.
During diffusion inference, the model reconstructs $W_{\mathrm{rec}}$
from deterministic 2-bit RTN conditions.
The resulting $W_{\mathrm{rec}}$ is computed once and reused unchanged
for both the 3-bit and 4-bit ReRound experiments.

For the SINQ experiments, we instead train separate diffusion models
for the 3-bit and 4-bit conditions.
At each precision, SR is used to generate the training conditions,
whereas deterministic RTN is used to generate the conditions during
diffusion inference.

\section{ReRound Settings}
\label{sec:supp_reround_ptq_details}

\paragraph{ReRound PTQ.}
For each target bit width, ReRound uses one GPU to map $W$ and
$W_{\mathrm{rec}}$ to the same quantized weight space, construct
candidate quantized integer weight matrices over
$\tau\in\mathcal{T}$, and select one candidate for each weight matrix
by matching the leading singular values of its dequantized weight
matrix to those of $W$.
We use separate candidate sets for the transformer linear layers and
the output logit layer (\texttt{lm\_head}).
Table~\ref{tab:reround_hyperparameters} reports these candidate sets
and the position-dependent tolerance parameter $\beta$.
The same settings are used for the 3-bit and 4-bit experiments.
A candidate with $\tau=0$ corresponds to retaining the RTN
assignments.
ReRound changes at most $1\%$ of the RTN quantized integer assignments
in each matrix and uses the same formulation across models.
All diffusion-based reconstruction and ReRound PTQ stages use only the
pretrained model's weights, without activation or text calibration
samples.
In Eq.~(13) of the main paper, we set $\delta=0.1$ for all models and
bit widths, corresponding to the central midpoint ambiguity region
$r\in[0.4,0.6]$.

\begin{table}[t]
\centering
\small
\setlength{\tabcolsep}{3.2pt}
\resizebox{0.8\columnwidth}{!}{
\begin{tabular}{@{}lccc@{}}
\toprule
Model
& $\mathcal{T}_{\mathrm{linear}}$
& $\mathcal{T}_{\mathrm{head}}$
& $\beta$ \\
\midrule
Gemma 2 2B
& $\{0, 0.15,\,0.25\}$
& $\{0, 0.15,\,0.25\}$
& 8 \\

Gemma 3 1B
& $\{0, 0.15,\,0.25\}$
& $\{0, 0.15,\,0.25\}$
& 64 \\

Qwen3 1.7B
& $\{0,\,0.15\}$
& $\{0\}$
& 128 \\

OLMo 2 1B
& $\{0, 0.15,\,0.25\}$
& $\{0, 0.15,\,0.25\}$
& 32 \\

SmolLM2 1.7B
& $\{0, 0.15,\,0.25\}$
& $\{0\}$
& 128 \\

Llama 3.2 1B
& $\{0, 0.15,\,0.25,\,0.35\}$
& $\{0, 0.15\}$
& 64 \\

Pythia 1.4B
& $\{0,\,0.15\}$
& $\{0\}$
& 8 \\

Phi-2 2.7B
& $\{0,\,0.15\}$
& $\{0\}$
& 64 \\

Falcon 3-1B-Base
& $\{0,\,0.15\}$
& $\{0\}$
& 8 \\
\bottomrule
\end{tabular}
}
\caption{\textbf{ReRound hyperparameters for each model.}
$\mathcal{T}_{\mathrm{linear}}$ is the tolerance candidate set used
for the transformer linear layers, and
$\mathcal{T}_{\mathrm{head}}$ is the set used for the output logit
layer.
The parameter $\beta$ controls the decrease of the
position-dependent tolerance outside the midpoint ambiguity region.
The same settings are used at 3 and 4 bits.}
\label{tab:reround_hyperparameters}
\end{table}

\paragraph{ReRound with SINQ parameters.}
For the experiments using SINQ quantization parameters on Qwen3 1.7B,
we follow SINQ's layer coverage and leave the output logit layer
(\texttt{lm\_head}) unquantized.
Table~\ref{tab:sinq_reround_hyperparameters} reports the tolerance
candidate sets used for the remaining transformer linear layers.
We use $\beta=8$ at both bit widths.
As before, $\tau=0$ corresponds to retaining the RTN assignments.

\begin{table}[t]
\centering
\small
\setlength{\tabcolsep}{6pt}
\begin{tabular}{@{}ccc@{}}
\toprule
Weight precision
& $\mathcal{T}_{\mathrm{linear}}$
& $\beta$ \\
\midrule
W4
& $\{0,\,0.15\}$
& 8 \\
W3
& $\{0, 0.15,\,0.25,\,0.35\}$
& 8 \\
\bottomrule
\end{tabular}
\caption{\textbf{ReRound hyperparameters with SINQ quantization
parameters on Qwen3 1.7B.}
The \texttt{lm\_head} is left unquantized following the SINQ layer
coverage.}
\label{tab:sinq_reround_hyperparameters}
\end{table}

\section{Additional PTQ Results}
\label{sec:supp_additional_ptq_results}

\subsection{RTN Improvements on Additional Models}
\label{sec:supp_rtn_improvements_across_models}

Table~\ref{tab:supp_additional_model_results} reports the complete
per-task results for Llama 3.2 1B~\cite{GrattafioriDJKD24},
Pythia 1.4B~\cite{BidermanSABHKPR23}, and
Phi-2 2.7B~\cite{JavaheripiB23}, together with results on
Falcon 3-1B-Base~\cite{Falcon3Team24}.
We compare ReRound with group-wise RTN under W4A16 and W3A16
quantization using the same scales, zero-points, group size $G=128$,
and layer coverage.
ReRound achieves a higher four-task average for every model at both
bit widths, with improvements of up to $1.6$ points.
Because the quantization setup remains unchanged, these gains result
solely from revising selected quantized integer assignments.

\begin{table}[t]
\centering
\small
\setlength{\tabcolsep}{3.2pt}
\newlength{\ArchAblationOneColWidth}
\setlength{\ArchAblationOneColWidth}{0.99\columnwidth}

\resizebox{\ArchAblationOneColWidth}{!}{
\begin{tabular}{@{}l c l ccccA@{}}
\toprule
Model & Bits & Method
& WinoGrande$\uparrow$ & PIQA$\uparrow$ & BoolQ$\uparrow$
& SIQA$\uparrow$ & Avg.$\uparrow$ \\
\midrule

\multirow{5}{*}{Llama 3.2 1B}
& 16 bits
& --
& 60.3 & 74.6 & 63.7 & 43.0 & 60.4 \\
\cmidrule(lr){2-8}
& \multirow{2}{*}{W4A16}
& RTN group-wise, $G=128$
& 61.1 & 72.4 & 47.1 & \textbf{41.8} & 55.6 \\
&
& \textbf{ReRound}, $G=128$
& \textbf{61.3} & \textbf{72.5} & \textbf{50.4} & 41.7
& \textbf{56.5} \\
\cmidrule(lr){2-8}
& \multirow{2}{*}{W3A16}
& RTN group-wise, $G=128$
& \textbf{53.6} & \textbf{63.0} & 50.1 & 37.7 & 51.1 \\
&
& \textbf{ReRound}, $G=128$
& 51.4 & 62.8 & \textbf{54.3} & \textbf{38.3}
& \textbf{51.7} \\

\midrule

\multirow{5}{*}{Falcon 3-1B-Base}
& 16 bits
& --
& 61.7 & 74.5 & 71.8 & 45.2 & 63.3 \\
\cmidrule(lr){2-8}
& \multirow{2}{*}{W4A16}
& RTN group-wise, $G=128$
& \textbf{59.6} & 73.8 & 70.6 & 44.4 & 62.1 \\
&
& \textbf{ReRound}, $G=128$
& \textbf{59.6} & \textbf{74.2} & \textbf{70.8}
& \textbf{45.0} & \textbf{62.4} \\
\cmidrule(lr){2-8}
& \multirow{2}{*}{W3A16}
& RTN group-wise, $G=128$
& 58.6 & 71.9 & 62.9 & \textbf{41.3} & 58.7 \\
&
& \textbf{ReRound}, $G=128$
& \textbf{59.0} & \textbf{72.4} & \textbf{63.1}
& 40.6 & \textbf{58.8} \\

\midrule

\multirow{5}{*}{Phi-2 2.7B}
& 16 bits
& --
& 76.2 & 78.6 & 83.1 & 55.5 & 73.4 \\
\cmidrule(lr){2-8}
& \multirow{2}{*}{W4A16}
& RTN group-wise, $G=128$
& 75.5 & \textbf{78.6} & \textbf{81.1} & 54.7 & 72.5 \\
&
& \textbf{ReRound}, $G=128$
& \textbf{76.5} & 78.2 & 80.7 & \textbf{55.6}
& \textbf{72.8} \\
\cmidrule(lr){2-8}
& \multirow{2}{*}{W3A16}
& RTN group-wise, $G=128$
& 71.7 & 77.6 & 64.7 & 51.3 & 66.3 \\
&
& \textbf{ReRound}, $G=128$
& \textbf{73.6} & \textbf{78.4} & \textbf{67.1}
& \textbf{52.3} & \textbf{67.9} \\

\midrule

\multirow{5}{*}{Pythia 1.4B}
& 16 bits
& --
& 57.5 & 71.1 &63.2 & 41.2 & 58.3 \\
\cmidrule(lr){2-8}
& \multirow{2}{*}{W4A16}
& RTN group-wise, $G=128$
& 55.9 & \textbf{70.2} & 61.6 & 40.2 & 57.0 \\
&
& \textbf{ReRound}, $G=128$
& \textbf{56.1} & 70.0 & \textbf{62.3}
& \textbf{41.1} & \textbf{57.4} \\
\cmidrule(lr){2-8}
& \multirow{2}{*}{W3A16}
& RTN group-wise, $G=128$
& 52.7 & \textbf{67.0} & 52.8 & 38.8 & 52.8 \\
&
& \textbf{ReRound}, $G=128$
& \textbf{53.0} & 66.7 & \textbf{55.9}
& \textbf{39.0} & \textbf{53.7} \\

\bottomrule
\end{tabular}
}

\caption{
\textbf{Additional results across LLM architectures.}
We compare ReRound with group-wise RTN on Llama 3.2 1B,
Falcon 3-1B-Base, Phi-2 2.7B, and Pythia 1.4B.
Both methods use the same asymmetric group-wise quantization setup
with group size $G=128$.
W4A16 and W3A16 denote 4-bit and 3-bit weight quantization with
16-bit activations, respectively.
Avg. reports the mean accuracy over WinoGrande, PIQA, BoolQ, and SIQA.
The better quantized result for each model and bit width is shown in
bold.
}
\label{tab:supp_additional_model_results}
\end{table}

\subsection{Comparison with AdaRound}
\label{sec:supp_adaround_comparison}

Table~\ref{tab:supp_adaround_reround_w4} compares AdaRound and ReRound
under W4A16 quantization on OLMo 2 1B~\cite{TeamWSGLABGHJ25},
SmolLM2 1.7B~\cite{AllalLBMPTMKLS25},
Llama 3.2 1B~\cite{GrattafioriDJKD24},
Falcon 3-1B-Base~\cite{Falcon3Team24},
Phi-2 2.7B~\cite{JavaheripiB23}, and
Qwen3 1.7B~\cite{YangLAYZHBGYHL25}.
Both methods make layer-wise rounding decisions.
AdaRound uses 128 calibration samples from C4~\cite{RaffelSRLNMZLL20},
while ReRound uses only the pretrained model weights.
ReRound achieves a higher four-task average than AdaRound on all six
models, with improvements of up to $1.9$ points.
Although AdaRound performs better on some individual tasks, ReRound
consistently provides higher average accuracy without activation or
text calibration data.

\begin{table}[t]
\centering
\small
\setlength{\tabcolsep}{3.6pt}
\newlength{\AdaRoundSuppTableWidth}
\setlength{\AdaRoundSuppTableWidth}{0.99\columnwidth}

\resizebox{\AdaRoundSuppTableWidth}{!}{
\begin{tabular}{@{}l l ccccA@{}}
\toprule
Model & Method
& WinoGrande$\uparrow$ & PIQA$\uparrow$ & BoolQ$\uparrow$ & SIQA$\uparrow$ & Avg.$\uparrow$ \\
\midrule

\multirow{2}{*}{OLMo 2 1B}
& AdaRound, $G=128$
& 63.4 & 74.9 & 48.1 & \textbf{43.7} & 57.5 \\
& \textbf{ReRound}, $G=128$
& \textbf{63.9} & \textbf{75.4} & \textbf{54.4} & \textbf{43.7} & \textbf{59.4} \\

\midrule

\multirow{2}{*}{SmolLM2 1.7B}
& AdaRound, $G=128$
& 62.4 & 76.0 & \textbf{69.4} & \textbf{42.7} & 62.6 \\
& \textbf{ReRound}, $G=128$
& \textbf{65.1} & \textbf{76.7} & 67.3 & 42.3 & \textbf{62.9} \\

\midrule

\multirow{2}{*}{Llama 3.2 1B}
& AdaRound, $G=128$
& 61.2 & \textbf{72.7} & 46.2 & \textbf{41.9} & 55.5 \\
& \textbf{ReRound}, $G=128$
& \textbf{61.3} & 72.5 & \textbf{50.4} & 41.7 & \textbf{56.5} \\

\midrule

\multirow{2}{*}{Falcon 3-1B-Base}
& AdaRound, $G=128$
& \textbf{60.1} & 74.0 & 70.2 & 44.4 & 62.2 \\
& \textbf{ReRound}, $G=128$
& 59.6 & \textbf{74.2} & \textbf{70.8} & \textbf{45.0} & \textbf{62.4} \\

\midrule

\multirow{2}{*}{Phi-2 2.7B}
& AdaRound, $G=128$
& 75.9 & 77.9 & \textbf{80.7} & 54.7 & 72.3 \\
& \textbf{ReRound}, $G=128$
& \textbf{76.5} & \textbf{78.2} & \textbf{80.7} & \textbf{55.6} & \textbf{72.8} \\

\midrule

\multirow{2}{*}{Qwen3 1.7B}
& AdaRound, $G=128$
& 58.1 & \textbf{70.4} & 77.2 & 42.0 & 61.9 \\
& \textbf{ReRound}, $G=128$
& \textbf{58.6} & 70.3 & \textbf{78.3} & \textbf{43.2} & \textbf{62.6} \\

\bottomrule
\end{tabular}
}

\caption{
Comparison with AdaRound under W4A16 quantization on OLMo 2 1B, SmolLM2 1.7B, Llama 3.2 1B, Falcon 3-1B-Base, Phi-2 2.7B, and Qwen3 1.7B.
AdaRound learns layer-wise rounding decisions using calibration data, while ReRound selects layer-wise rounding decisions without calibration data.
Both methods use the same group-wise scale, zero-point, and group size $G=128$.
Avg. is over WinoGrande, PIQA, BoolQ, and SIQA. Higher is better.
}
\label{tab:supp_adaround_reround_w4}
\end{table}

\section{Baseline Implementation Details}
\label{sec:supp_baseline_details}
Unless otherwise stated, all reimplemented baselines use the model
checkpoints and evaluation protocol described in
Section~\ref{sec:supp_models_evaluation}.
The transformer linear layers and output logit layer are quantized in
the main comparisons, and all group-wise methods use group size
$G=128$.
\paragraph{RTN.}
We implement uniform asymmetric RTN using the minimum and maximum
weight values for computing the scale and zero-point.
Channel-wise RTN computes one pair of quantization parameters per
output channel, whereas group-wise RTN computes one pair for every
group of 128 consecutive weights along the input dimension.
\paragraph{HQQ.}
HQQ version 0.2.8.post1 is applied with the PyTorch backend,
quantization axis 1, and group size $G=128$.
\paragraph{BNB FP4.}
BitsAndBytes version 0.49.2 is applied with FP4 weight quantization and
BF16 computation.
BNB FP4 is evaluated only under W4A16.
\paragraph{Hadamard-transformed RTN.}
We apply a normalized Hadamard transform independently to groups of
128 consecutive weights along the input-feature dimension, perform
asymmetric RTN in the transformed domain, and then apply the inverse
transform.
\paragraph{GPTQ.}
GPTQ is applied to all transformer linear layers using asymmetric
group-wise quantization with group size $G=128$.
We use 128 calibration sequences of length 2048 from the C4 training
split~\cite{RaffelSRLNMZLL20} and enable activation ordering and
true-sequential quantization.
The output logit layer uses asymmetric group-wise RTN with $G=128$.
\paragraph{AdaRound.}
AdaRound uses asymmetric group-wise quantization with group size
$G=128$ and 128 calibration sequences of length 2048 from the C4
training split~\cite{RaffelSRLNMZLL20}.
Rounding variables are optimized for 200 iterations using up to 1024
calibration tokens per weight matrix, a learning rate of $10^{-2}$,
and a regularization weight of $0.01$.
To accommodate the large output logit layer, we divide it into blocks
covering 1024 vocabulary outputs and optimize each block for 50
iterations.
\paragraph{SignRound.}
We use AutoRound version 0.13.1 with its default optimization settings,
symmetric group-wise weight quantization, and group size $G=128$.
The output logit layer is also quantized using AutoRound's built-in
\texttt{quant\_lm\_head} procedure.

\section{Models and Evaluation Protocol}
\label{sec:supp_models_evaluation}

\paragraph{Model checkpoints.}
Table~\ref{tab:model_checkpoints} lists the exact Hugging Face
checkpoints used in our experiments.

\begin{table}[t]
\centering
\small
\setlength{\tabcolsep}{4pt}
\resizebox{0.92\columnwidth}{!}{
\begin{tabular}{@{}ll@{}}
\toprule
Model & Hugging Face checkpoint \\
\midrule
Gemma 2 2B
& \texttt{google/gemma-2-2b} \\
Gemma 3 1B
& \texttt{google/gemma-3-1b-pt} \\
Qwen3 1.7B
& \texttt{Qwen/Qwen3-1.7B} \\
OLMo 2 1B
& \texttt{allenai/OLMo-2-0425-1B} \\
SmolLM2 1.7B
& \texttt{HuggingFaceTB/SmolLM2-1.7B} \\
Llama 3.2 1B
& \texttt{meta-llama/Llama-3.2-1B} \\
Pythia 1.4B
& \texttt{EleutherAI/pythia-1.4b} \\
Phi-2 2.7B
& \texttt{microsoft/phi-2} \\
Falcon 3-1B-Base
& \texttt{tiiuae/Falcon3-1B-Base} \\
\bottomrule
\end{tabular}
}
\caption{\textbf{Exact model checkpoints used in our experiments.}}
\label{tab:model_checkpoints}
\end{table}

\paragraph{Accuracy evaluation.}
All accuracy evaluations are zero-shot and use the standard task
configurations in LM Evaluation Harness v0.4.11.
We report accuracy (\texttt{acc}) for WinoGrande, BoolQ, and SIQA, and
normalized accuracy (\texttt{acc\_norm}) for PIQA.
Activations remain in 16-bit precision for all weight-quantized
experiments.
Across all methods, we use random seed 0 and set the NumPy and PyTorch
seeds to 1234.

\begin{figure*}[t]
    \centering
    \includegraphics[width=0.7\textwidth]{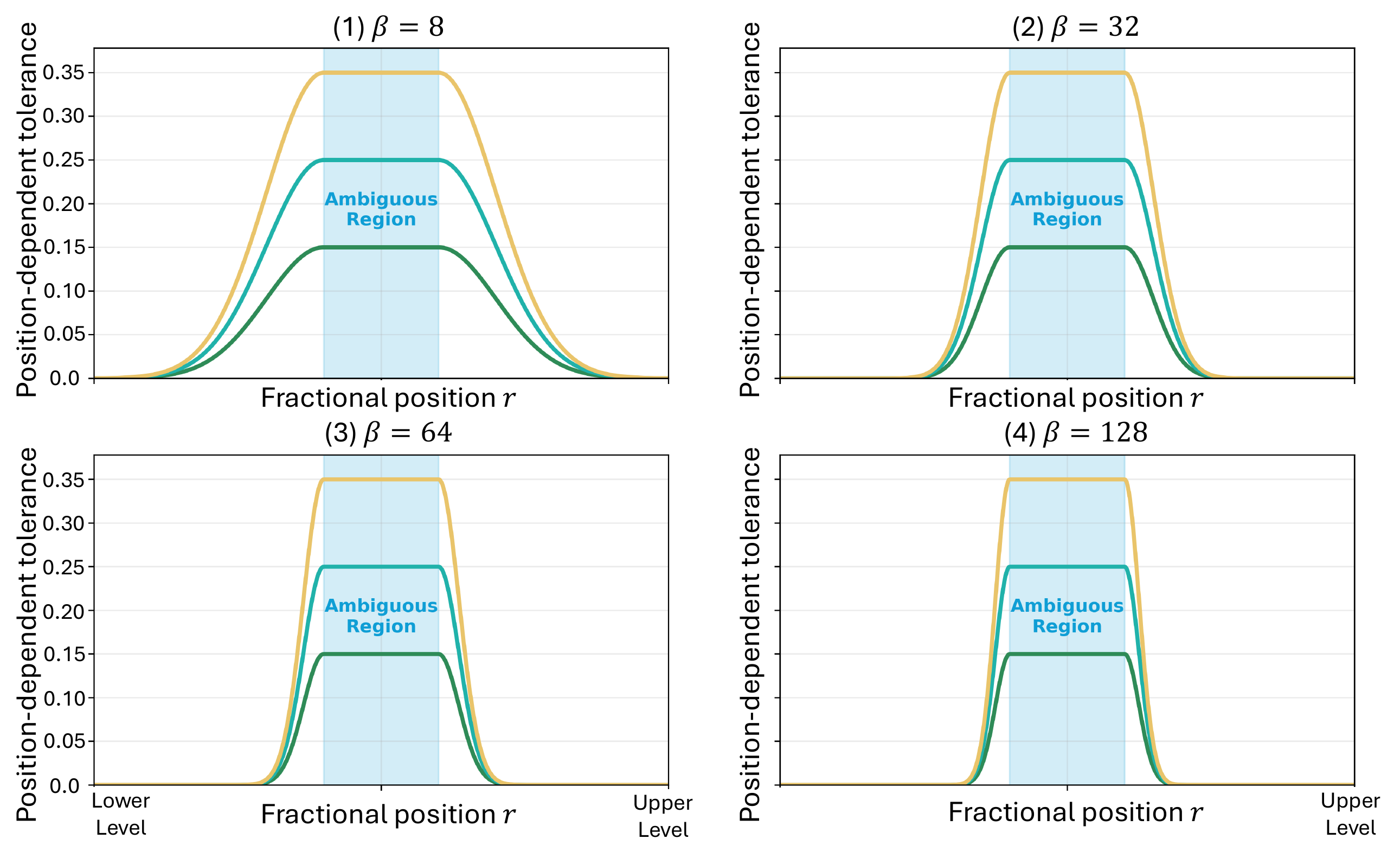}
    \caption{\textbf{Visualization of the position-dependent tolerance
    metric.}
    We plot $\tau_{\mathrm{pos}}(r;\tau)$ for the illustrative tolerance
    parameters $\tau\in\{0.15,0.25,0.35\}$ with
    (1) $\beta=8$, (2) $\beta=32$, (3) $\beta=64$, and
    (4) $\beta=128$.
    The \textcolor[HTML]{0F9ED5}{blue shaded interval} denotes the midpoint ambiguity region
    $r\in[0.4,0.6]$, defined by $\delta=0.1$, where
    $\tau_{\mathrm{pos}}(r;\tau)=\tau$.
    Outside this region, the tolerance metric decreases toward either
    adjacent quantized integer.
    Larger values of $\beta$ produce sharper decreases at the
    boundaries of the midpoint ambiguity region.
    }
    \label{fig:position_dependent_tolerance}
\end{figure*}

\section{Visualization of the Position-Dependent Tolerance Metric}

Figure~\ref{fig:position_dependent_tolerance} visualizes the
position-dependent tolerance metric
$\tau_{\mathrm{pos}}(r;\tau)$ for different values of $\beta$.
For illustration, we show the tolerance parameters
$\tau\in\{0.15,0.25,0.35\}$.
With $\delta=0.1$, the midpoint ambiguity region is
$|r-0.5|\leq\delta$, or equivalently $r\in[0.4,0.6]$.
Within this region,
$\tau_{\mathrm{pos}}(r;\tau)=\tau$.
Outside the midpoint ambiguity region, $\beta$ controls how quickly the tolerance
metric decreases toward either adjacent quantized integer.
A smaller $\beta$ produces a more gradual decrease, whereas a larger
$\beta$ produces a sharper decrease at the boundaries of the midpoint
region.

\begin{figure*}[t]
    \centering
    \begin{minipage}[t]{0.49\textwidth}
        \centering
        \includegraphics[width=\linewidth]
        {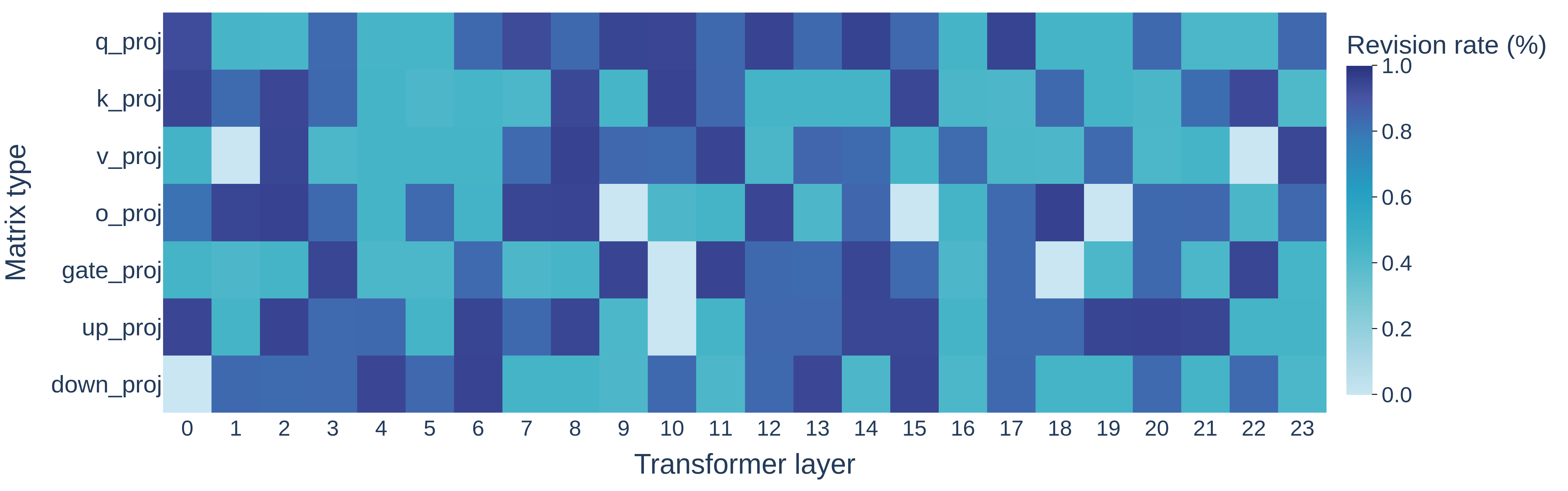}
        \vspace{-2mm}
        \textbf{(a) W4}
    \end{minipage}
    \hfill
    \begin{minipage}[t]{0.49\textwidth}
        \centering
        \includegraphics[width=\linewidth]
        {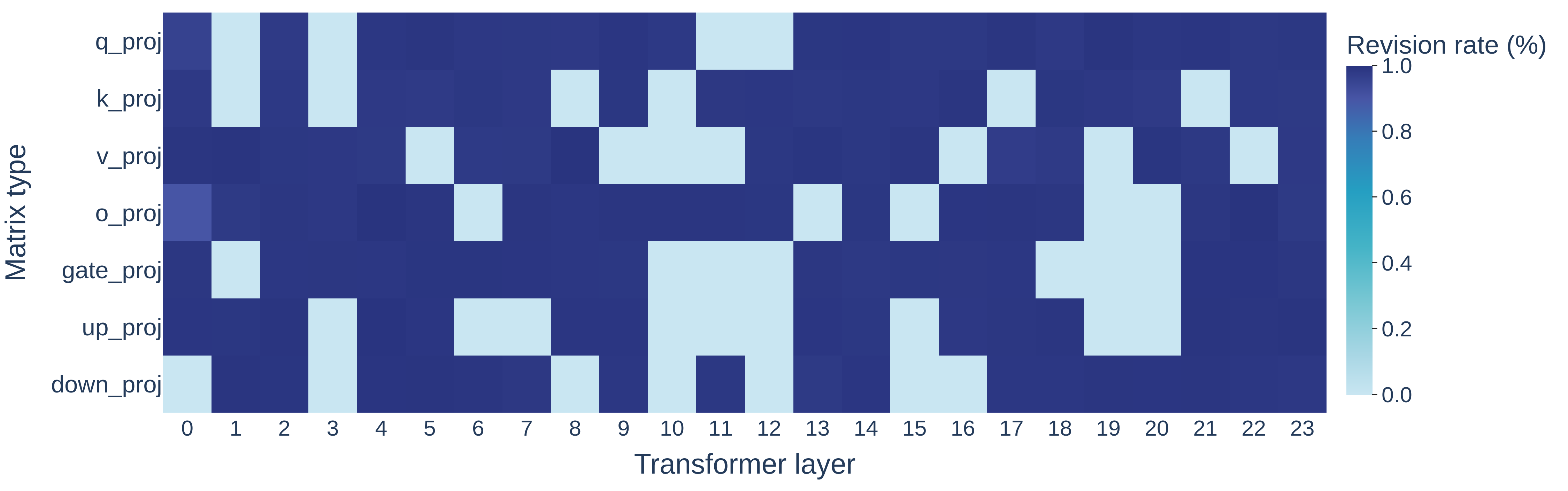}
        \vspace{-2mm}
        \textbf{(b) W3}
    \end{minipage}
    \caption{
    \textbf{Matrix-wise RTN revision rates for SmolLM2 1.7B.}
    Each cell shows the percentage of RTN assignments revised by ReRound
    for one weight matrix; lighter and darker colors indicate lower and
    higher revision rates, respectively.
    The revisions are distributed nonuniformly across network depth and
    matrix type, and the W3 and W4 patterns differ despite reusing the same
    reconstructed weights.
    This shows that ReRound does not derive a fixed set of revisions from
    the reconstruction alone.
    Instead, each matrix is evaluated under its bit-specific quantization
    grid and spectral preservation objective.
    The same color scale is used in both panels.
    }
    \label{fig:smollm2_revision_heatmaps}
\end{figure*}

\section{Matrix-Wise Revision of RTN Assignments}
\label{sec:matrix_wise_revision}

\begin{figure*}[t]
    \centering
    \begin{minipage}[t]{0.49\textwidth}
        \centering
        \includegraphics[width=\linewidth]
        {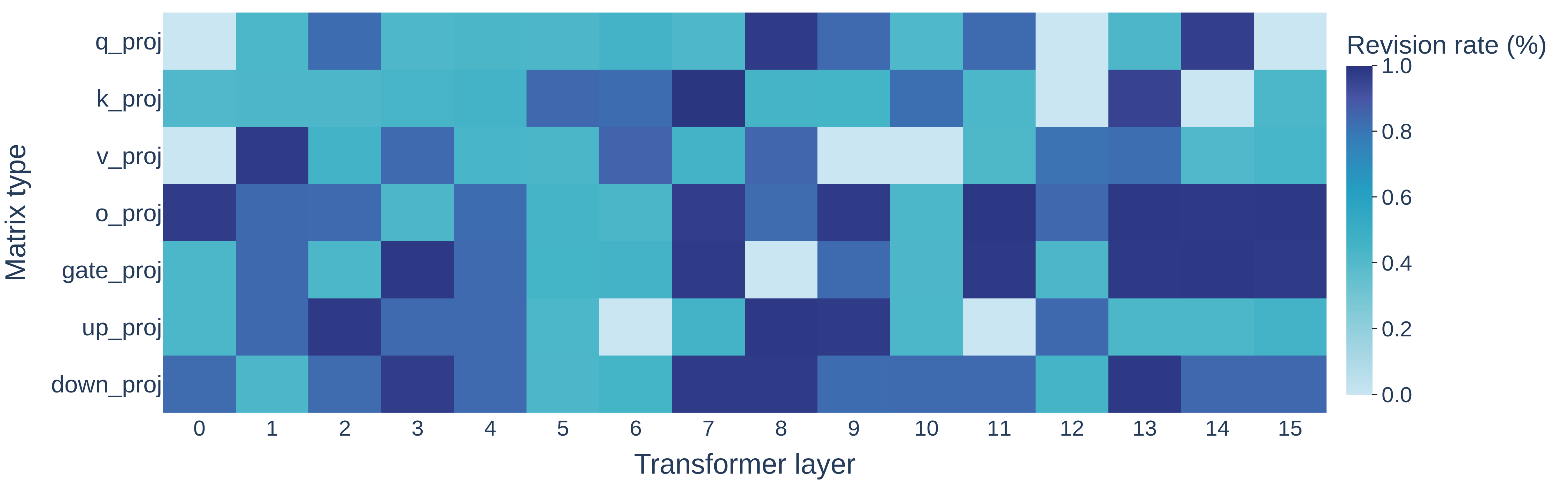}
        \vspace{-2mm}
        \textbf{(a) W4}
    \end{minipage}
    \hfill
    \begin{minipage}[t]{0.49\textwidth}
        \centering
        \includegraphics[width=\linewidth]
        {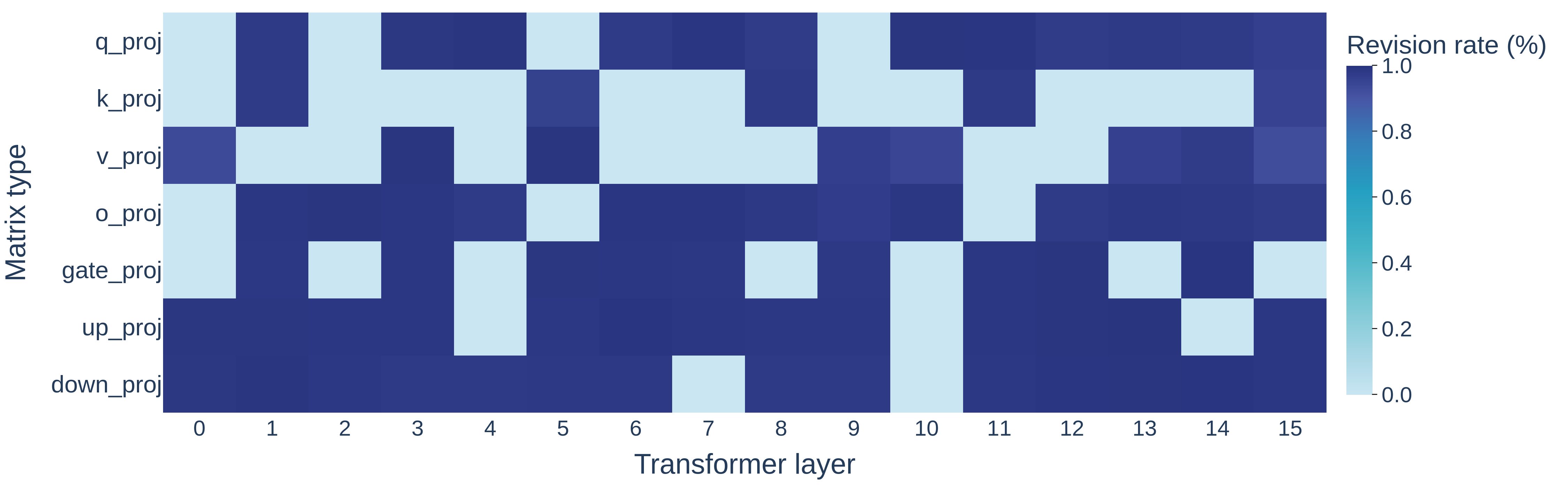}
        \vspace{-2mm}
        \textbf{(b) W3}
    \end{minipage}
    \caption{
    \textbf{Matrix-wise RTN revision rates for Llama 3.2 1B.}
    Each cell shows the percentage of RTN assignments revised by ReRound
    for one weight matrix; lighter and darker colors indicate lower and
    higher revision rates, respectively.
    Although W3 and W4 use the same reconstructed weights, they produce
    different matrix-wise revision patterns.
    ReRound evaluates the rounding candidates separately for each matrix
    and bit width, and selects the candidate that best preserves the
    leading singular values of the matrix.
    The same color scale is used in both panels.
    }
    \label{fig:llama32_revision_heatmaps}
\end{figure*}

For each weight matrix, we report the percentage of RTN assignments
revised by ReRound.
Figures~\ref{fig:smollm2_revision_heatmaps} and
\ref{fig:llama32_revision_heatmaps} visualize these revision rates
across layers and matrix types under W3 and W4 quantization.

The revision rates vary across matrices and bit widths.
Although W3 and W4 reuse the same reconstructed weights, the same
matrix can receive different revision rates at the two bit widths.
This is because ReRound evaluates the rounding candidates under each
target quantization grid and performs spectral selection separately
for every matrix.
Therefore, the amount of revision depends on both the weight matrix
and the target bit width, rather than on the reconstructed weights
alone.

\section{Continuous Reconstruction for Low-Bit Rounding}
\label{sec:supp_continuous_reconstruction}

ReRound uses a DDPM-based conditional diffusion model
~\cite{HoJA20,SahariaHCSFN22} to produce continuous reconstructed
weights from low-bit weight observations.
The reconstructed weights are not deployed as model parameters.
They provide an additional signal for reconsidering
midpoint-ambiguous RTN assignments.

\subsection{Low-Bit Conditioning}

Let $P$ denote a normalized patch cropped from a full-precision weight
matrix $W$.
During diffusion training, its low-bit condition is obtained by
applying group-wise stochastic rounding (SR):
\begin{equation}
P_{\mathrm{low}}
=
Q_{\mathrm{sr}}(P;\Delta_c,z_c),
\label{eq:supp_training_condition}
\end{equation}
where $Q_{\mathrm{sr}}$ denotes the group-wise SR operation.
The parameters $\Delta_c$ and $z_c$ are the scale and zero-point used
to construct the 2-bit diffusion condition.
The subscript $c$ distinguishes these conditioning parameters from
the target quantization parameters $\Delta$ and $z$ used later for
W3 or W4 rounding.

During diffusion inference, the condition is instead generated
deterministically using RTN.
For a patch $P_k$,
\begin{equation}
P_{\mathrm{cond},k}
=
Q_{\mathrm{rtn}}(P_k;\Delta_c,z_c).
\label{eq:supp_inference_condition}
\end{equation}

For fixed $\Delta_c$ and $z_c$, deterministic low-bit quantization is
many-to-one.
Distinct continuous patches may produce the same RTN condition:
\begin{equation}
P^{(1)}\neq P^{(2)},
\qquad
Q_{\mathrm{rtn}}(P^{(1)};\Delta_c,z_c)
=
Q_{\mathrm{rtn}}(P^{(2)};\Delta_c,z_c).
\label{eq:supp_many_to_one_condition}
\end{equation}
The low-bit condition therefore does not uniquely identify the
original full-precision patch.

The diffusion model represents a conditional distribution over
full-precision patches given their low-bit conditions
~\cite{SahariaHCSFN22}:
\begin{equation}
p_{\theta}(P\mid P_{\mathrm{low}})
\approx
p_{\mathrm{data}}(P\mid P_{\mathrm{low}}).
\label{eq:supp_conditional_patch_distribution}
\end{equation}
Because the entries of a patch are reconstructed jointly, each
reconstructed entry can depend on both its own low-bit observation
and the surrounding entries in the patch.

\subsection{Conditional Denoising Relation}

The following relations are defined under the joint training
distribution of the full-precision patch $P$, its stochastic condition
$P_{\mathrm{low}}$, the diffusion timestep $t$, and Gaussian noise
$\epsilon$.
Define
\begin{equation}
\alpha_t=\sqrt{\bar{\alpha}_t},
\qquad
\sigma_t=\sqrt{1-\bar{\alpha}_t}.
\label{eq:supp_diffusion_coefficients}
\end{equation}
Following the DDPM forward process~\cite{HoJA20}, Gaussian noise is
added as
\begin{equation}
P_t
=
\alpha_t P+\sigma_t\epsilon,
\qquad
\epsilon\sim\mathcal{N}(0,I),
\label{eq:supp_forward_diffusion}
\end{equation}
where $\epsilon$ is sampled independently of
$(P,P_{\mathrm{low}})$.
For a fixed $P$ and timestep $t$, the forward transition is
\begin{equation}
q_t(P_t\mid P)
=
\mathcal{N}
\left(
P_t;\alpha_tP,\sigma_t^2I
\right).
\label{eq:supp_forward_transition}
\end{equation}

Let $p_t(P_t\mid P_{\mathrm{low}})$ denote the conditional marginal
distribution of the noisy patch.
Differentiating this marginal with respect to $P_t$ gives
\begin{align}
\nabla_{P_t}
\log p_t(P_t\mid P_{\mathrm{low}})
&=
\mathbb{E}
\left[
\nabla_{P_t}
\log q_t(P_t\mid P)
\;\middle|\;
P_t,P_{\mathrm{low}}
\right]
\nonumber\\
&=
-\frac{
P_t-
\alpha_t
\mathbb{E}[P\mid P_t,P_{\mathrm{low}}]
}{
\sigma_t^2
}.
\label{eq:supp_conditional_score_expanded}
\end{align}
The second equality follows from
\begin{equation}
\nabla_{P_t}\log q_t(P_t\mid P)
=
-\frac{P_t-\alpha_tP}{\sigma_t^2}.
\end{equation}

Rearranging
Eq.~\eqref{eq:supp_conditional_score_expanded} gives the conditional
Gaussian form of Tweedie's formula~\cite{Efron11}:
\begin{equation}
\mathbb{E}[P\mid P_t,P_{\mathrm{low}}]
=
\frac{
P_t+
\sigma_t^2
\nabla_{P_t}
\log p_t(P_t\mid P_{\mathrm{low}})
}{
\alpha_t
}.
\label{eq:supp_conditional_tweedie}
\end{equation}

We can also express the conditional score through the added noise.
Taking the conditional expectation of
Eq.~\eqref{eq:supp_forward_diffusion} given
$(P_t,t,P_{\mathrm{low}})$ gives
\begin{equation}
P_t
=
\alpha_t
\mathbb{E}[P\mid P_t,t,P_{\mathrm{low}}]
+
\sigma_t
\mathbb{E}[\epsilon\mid P_t,t,P_{\mathrm{low}}].
\label{eq:supp_conditioned_forward}
\end{equation}
Solving for the conditional noise mean yields
\begin{equation}
\mathbb{E}[\epsilon\mid P_t,t,P_{\mathrm{low}}]
=
\frac{
P_t-
\alpha_t
\mathbb{E}[P\mid P_t,t,P_{\mathrm{low}}]
}{
\sigma_t
}.
\label{eq:supp_conditional_noise_mean}
\end{equation}
Combining
Eqs.~\eqref{eq:supp_conditional_score_expanded} and
\eqref{eq:supp_conditional_noise_mean} gives
\begin{equation}
\nabla_{P_t}
\log p_t(P_t\mid P_{\mathrm{low}})
=
-\frac{1}{\sigma_t}
\mathbb{E}[\epsilon\mid P_t,t,P_{\mathrm{low}}].
\label{eq:supp_conditional_score}
\end{equation}
The conditional diffusion model minimizes
\begin{equation}
\mathcal{L}_{\mathrm{diff}}
=
\mathbb{E}_{P,P_{\mathrm{low}},t,\epsilon}
\left[
\left\|
\epsilon-
\epsilon_{\theta}(P_t,t,P_{\mathrm{low}})
\right\|_2^2
\right].
\label{eq:supp_noise_prediction_objective}
\end{equation}
For a fixed input $(P_t,t,P_{\mathrm{low}})$, let
\begin{equation}
\mu_{\epsilon}
=
\mathbb{E}[\epsilon\mid P_t,t,P_{\mathrm{low}}].
\end{equation}
For any noise predictor $f(P_t,t,P_{\mathrm{low}})$, the conditional
squared error decomposes as
\begin{align}
&
\mathbb{E}
\left[
\left\|
\epsilon-f(P_t,t,P_{\mathrm{low}})
\right\|_2^2
\;\middle|\;
P_t,t,P_{\mathrm{low}}
\right]
\nonumber\\
&=
\mathbb{E}
\left[
\left\|
\epsilon-\mu_{\epsilon}
\right\|_2^2
\;\middle|\;
P_t,t,P_{\mathrm{low}}
\right]
+
\left\|
f(P_t,t,P_{\mathrm{low}})
-
\mu_{\epsilon}
\right\|_2^2.
\label{eq:supp_squared_error_decomposition}
\end{align}
The first term does not depend on $f$, and the second term is minimized
when
\begin{equation}
f(P_t,t,P_{\mathrm{low}})
=
\mu_{\epsilon}.
\end{equation}
Therefore, the ideal noise predictor satisfies
\begin{equation}
\epsilon_{\theta}^{\star}
(P_t,t,P_{\mathrm{low}})
=
\mathbb{E}
[\epsilon\mid P_t,t,P_{\mathrm{low}}].
\label{eq:supp_ideal_noise_predictor}
\end{equation}
Substituting
Eq.~\eqref{eq:supp_ideal_noise_predictor} into
Eq.~\eqref{eq:supp_conditional_noise_mean} gives
\begin{equation}
\frac{
P_t-
\sigma_t
\epsilon_{\theta}^{\star}
(P_t,t,P_{\mathrm{low}})
}{
\alpha_t
}
=
\mathbb{E}
[P\mid P_t,t,P_{\mathrm{low}}].
\label{eq:supp_conditional_mean}
\end{equation}

Thus, for the ideal noise predictor, the clean estimate associated
with one denoising step equals the conditional mean of the
full-precision patch under Gaussian corruption
~\cite{Efron11,HoJA20}.
In practice, the learned denoiser approximates this predictor.
In addition, the relation above is defined using the stochastic
training condition $P_{\mathrm{low}}$, whereas reverse diffusion is
conditioned on the deterministic RTN patch $P_{\mathrm{cond}}$ during
inference.
The final multi-step output is therefore used as a continuous
reconstruction, not as an exact recovery of the original patch or as
an exact conditional mean.

\subsection{Midpoint Distance of Accepted Revisions}

Consider a target W3 or W4 quantization group with fixed scale
$\Delta$ and zero-point $z$.
For corresponding entries $w$ and $w_{\mathrm{rec}}$ of the
full-precision and reconstructed matrices, define
\begin{equation}
\tilde{w}
=
\frac{w}{\Delta}+z
=
\ell+r,
\qquad
\tilde{w}_{\mathrm{rec}}
=
\frac{w_{\mathrm{rec}}}{\Delta}+z,
\label{eq:supp_quantized_coordinates}
\end{equation}
where
\begin{equation}
\ell=\lfloor\tilde{w}\rfloor,
\qquad
u=\ell+1,
\qquad
r\in[0,1).
\end{equation}
The RTN assignment, its opposite adjacent assignment, and the
reconstruction-based assignment are
\begin{equation}
\begin{aligned}
q_{\mathrm{rtn}}
&=
\operatorname{round}(\tilde{w}),
&
q_{\mathrm{alt}}
&=
\ell+u-q_{\mathrm{rtn}},
\\
q_{\mathrm{rec}}
&=
\operatorname{round}(\tilde{w}_{\mathrm{rec}}).
\end{aligned}
\label{eq:supp_reconstruction_assignments}
\end{equation}
ReRound considers changing RTN only when
\begin{equation}
q_{\mathrm{rec}}=q_{\mathrm{alt}}.
\label{eq:supp_opposite_assignment}
\end{equation}
The acceptance threshold depends on the fractional position $r$.
Define
\begin{equation}
d(r)=\min\{r,1-r\},
\label{eq:supp_distance_to_integer}
\end{equation}
which equals $0.5$ at the midpoint and decreases toward either
adjacent quantized integer.

For tolerance parameter $\tau$, ReRound uses
\begin{equation}
\tau_{\mathrm{pos}}(r;\tau)
=
\tau
\exp
\left[
-\beta
\left(
\frac{
[0.5-\delta-d(r)]_{+}
}{
0.5-\delta
}
\right)^2
\right],
\label{eq:supp_position_dependent_tolerance}
\end{equation}
where $[x]_{+}=\max\{x,0\}$.
The parameter $\delta$ defines the central midpoint ambiguity region
$|r-0.5|\leq\delta$, in which
$\tau_{\mathrm{pos}}(r;\tau)=\tau$.
Outside this region, the tolerance decreases continuously according to
$\beta$; it is not set to zero.

The reconstruction deviation in quantization-step units is
\begin{equation}
\rho(w,w_{\mathrm{rec}})
=
\left|
\tilde{w}_{\mathrm{rec}}-\tilde{w}
\right|
=
\frac{|w_{\mathrm{rec}}-w|}{|\Delta|}.
\label{eq:supp_reconstruction_deviation}
\end{equation}
A revision is accepted only when
\begin{equation}
q_{\mathrm{rec}}=q_{\mathrm{alt}}
\qquad\text{and}\qquad
\rho(w,w_{\mathrm{rec}})
\leq
\tau_{\mathrm{pos}}(r;\tau).
\label{eq:supp_revision_conditions}
\end{equation}
The distance from the original quantized weight to the midpoint
boundary is
\begin{equation}
\gamma(r)
=
\left|r-\frac{1}{2}\right|.
\label{eq:supp_midpoint_distance}
\end{equation}
For the reconstructed assignment to equal the opposite adjacent
integer, $\tilde{w}_{\mathrm{rec}}$ must reach the other side of the
midpoint boundary.
Its displacement must therefore be at least $\gamma(r)$.
Every accepted revision consequently satisfies
\begin{equation}
\left|r-\frac{1}{2}\right|
\leq
\rho(w,w_{\mathrm{rec}})
\leq
\tau_{\mathrm{pos}}(r;\tau)
\leq
\tau.
\label{eq:supp_revision_distance}
\end{equation}

This relation follows directly from the realized reconstruction and
the ReRound acceptance conditions.
It does not depend on the denoiser being ideal.
An RTN assignment can be revised only when the quantized weight's
distance from the midpoint is no larger than the effective tolerance
at that position.

\subsection{Scalar Error of an Accepted Revision}

The squared errors of assigning $\tilde{w}=\ell+r$ to the lower and
upper adjacent integers are
\begin{equation}
e_{\ell}(r)=\Delta^2r^2,
\qquad
e_u(r)=\Delta^2(1-r)^2.
\label{eq:supp_adjacent_errors}
\end{equation}
Switching from RTN to the opposite adjacent assignment adds
\begin{equation}
\begin{aligned}
c_{\mathrm{flip}}(r)
&=
|e_u(r)-e_{\ell}(r)|
\\
&=
\Delta^2|1-2r|
\\
&=
2\Delta^2
\left|r-\frac{1}{2}\right|.
\end{aligned}
\label{eq:supp_flip_cost}
\end{equation}
Using Eq.~\eqref{eq:supp_revision_distance}, an accepted revision
satisfies
\begin{equation}
c_{\mathrm{flip}}(r)
\leq
2\Delta^2\tau_{\mathrm{pos}}(r;\tau)
\leq
2\Delta^2\tau.
\label{eq:supp_flip_cost_bound}
\end{equation}

The tolerance rule therefore bounds the additional scalar squared
quantization error of each accepted revision relative to RTN.
This is a scalar weight-space relation and does not characterize
changes in model outputs or downstream task accuracy.

\subsection{Spectral Comparison with RTN}

For each tolerance parameter $\tau\in\mathcal{T}$, applying the
ReRound rule element-wise produces a quantized integer matrix
$Q_{\tau}$.
Its dequantized matrix is
\begin{equation}
W_{\tau}
=
\operatorname{Dequant}(Q_{\tau};\Delta,z),
\label{eq:supp_candidate_dequantization}
\end{equation}
where $\Delta$ and $z$ are the fixed target quantization parameters.
The candidate $\tau=0$ retains all RTN assignments:
\begin{equation}
Q_0=Q_{\mathrm{rtn}},
\qquad
W_0=W_{\mathrm{rtn}}.
\label{eq:supp_rtn_candidate}
\end{equation}
For each candidate, ReRound measures
\begin{equation}
\mathcal{D}_{\mathrm{spec}}(\tau)
=
\frac{
\left\|
\sigma_{1:k}(W_{\tau})
-
\sigma_{1:k}(W)
\right\|_2
}{
\left\|
\sigma_{1:k}(W)
\right\|_2+\epsilon
},
\label{eq:supp_spectral_discrepancy}
\end{equation}
where $\sigma_{1:k}(\cdot)$ denotes the leading $k$ singular values,
and $\epsilon>0$ provides numerical stability.
For a matrix evaluated by the candidate search, ReRound selects
\begin{equation}
\tau^{\star}
=
\arg\min_{\tau\in\mathcal{T}}
\mathcal{D}_{\mathrm{spec}}(\tau).
\label{eq:supp_spectral_selection}
\end{equation}
Because $\tau=0$ is included in $\mathcal{T}$, RTN is one of the
candidates in this comparison.
It follows directly from the selection rule that
\begin{equation}
\mathcal{D}_{\mathrm{spec}}(\tau^{\star})
\leq
\mathcal{D}_{\mathrm{spec}}(0).
\label{eq:supp_spectral_comparison}
\end{equation}
The selected candidate therefore has no larger discrepancy than RTN
under the spectral criterion used for selection.
This inequality applies only to the spectral discrepancy used for
candidate selection.

The conditional model produces the continuous reconstruction used to
propose alternative assignments.
The position-dependent tolerance determines which proposals are
accepted, while spectral selection compares the resulting matrices
with the unchanged RTN candidate.

\begin{algorithm}[t]
\caption{Conditional Diffusion Weight Reconstruction}
\label{alg:weight_reconstruction}
\begin{algorithmic}[1]
\REQUIRE Full-precision weight matrices
$\{W^{(j)}\}_{j=1}^{M}$;
diffusion schedule $\{\bar{\alpha}_t\}_{t=1}^{T}$
\ENSURE Reconstructed weight matrices
$\{W_{\mathrm{rec}}^{(j)}\}_{j=1}^{M}$

\STATE \textbf{Training}
\FOR{each training iteration}
    \STATE Sample a matrix $W^{(j)}$, randomly crop a
    $64\times64$ region, and normalize it to obtain $P$
    \STATE Compute its 2-bit conditioning parameters
    $(\Delta_c,z_c)$
    \STATE
    $P_{\mathrm{low}}
    \leftarrow
    Q_{\mathrm{sr}}(P;\Delta_c,z_c)$
    \STATE Sample
    $t\sim\operatorname{Uniform}\{1,\ldots,T\}$ and
    $\epsilon\sim\mathcal{N}(0,I)$
    \STATE
    $P_t\leftarrow
    \sqrt{\bar{\alpha}_t}P+
    \sqrt{1-\bar{\alpha}_t}\epsilon$
    \STATE Update $\theta$ using
    \[
    \left\|
    \epsilon-
    \epsilon_{\theta}(P_t,t,P_{\mathrm{low}})
    \right\|_2^2
    \]
\ENDFOR

\STATE \textbf{Reconstruction}
\FOR{$j=1,\ldots,M$}
    \STATE Extract and normalize patches
    $\{P_k^{(j)}\}_{k=1}^{K_j}$ from $W^{(j)}$
    \FOR{$k=1,\ldots,K_j$}
        \STATE Compute the 2-bit conditioning parameters
        $(\Delta_c,z_c)$ for $P_k^{(j)}$
        \STATE
        $P_{\mathrm{cond},k}^{(j)}
        \leftarrow
        Q_{\mathrm{rtn}}
        (P_k^{(j)};\Delta_c,z_c)$
        \STATE Starting from $P_T\sim\mathcal{N}(0,I)$, apply
        conditional reverse diffusion to obtain
        $P_{\mathrm{rec},k}^{(j)}$
        \STATE Denormalize $P_{\mathrm{rec},k}^{(j)}$ and return it
        to its original location
    \ENDFOR
    \STATE Assemble the reconstructed patches into
    $W_{\mathrm{rec}}^{(j)}$
\ENDFOR

\RETURN $\{W_{\mathrm{rec}}^{(j)}\}_{j=1}^{M}$
\end{algorithmic}
\end{algorithm}

\begin{algorithm}[t]
\caption{ReRound for One Weight Matrix}
\label{alg:reround}
\begin{algorithmic}[1]
\REQUIRE Full-precision matrix $W$; reconstructed matrix
$W_{\mathrm{rec}}$; fixed target parameters $(\Delta,z)$;
integer range $[q_{\min},q_{\max}]$; candidate set
$\mathcal{T}$ with $0\in\mathcal{T}$; $\delta$, $\beta$;
maximum revision ratio $\eta$; spectral rank $k$
\ENSURE Selected quantized integer matrix $Q^\star$

\STATE Map both matrices to the target quantized space:
\[
\widetilde{W}\leftarrow W/\Delta+z,
\qquad
\widetilde{W}_{\mathrm{rec}}\leftarrow W_{\mathrm{rec}}/\Delta+z
\]

\STATE Compute
\[
L\leftarrow\lfloor\widetilde{W}\rfloor,\quad
R\leftarrow\widetilde{W}-L,\quad
D\leftarrow\min\{R,1-R\}
\]

\STATE Compute the rounding assignments:
\[
\begin{aligned}
Q_{\mathrm{rtn}}
&\leftarrow
\operatorname{clip}
(\operatorname{round}(\widetilde{W}),q_{\min},q_{\max}),\\
Q_{\mathrm{alt}}
&\leftarrow
\operatorname{clip}
(2L+1-Q_{\mathrm{rtn}},q_{\min},q_{\max}),\\
Q_{\mathrm{rec}}
&\leftarrow
\operatorname{clip}
(\operatorname{round}(\widetilde{W}_{\mathrm{rec}}),
q_{\min},q_{\max})
\end{aligned}
\]

\STATE
$\rho\leftarrow
|\widetilde{W}_{\mathrm{rec}}-\widetilde{W}|$,
\quad
$Q_0\leftarrow Q_{\mathrm{rtn}}$,
\quad
$\mathcal{T}_{\mathrm{valid}}\leftarrow\{0\}$

\FOR{each $\tau\in\mathcal{T}\setminus\{0\}$}
    \STATE
    $\displaystyle
    T_\tau\leftarrow
    \tau\exp\!\left[
    -\beta
    \left(
    \frac{[0.5-\delta-D]_+}{0.5-\delta}
    \right)^2
    \right]$

    \STATE
    $M_\tau\leftarrow
    (Q_{\mathrm{rec}}=Q_{\mathrm{alt}})
    \land(\rho\leq T_\tau)$

    \STATE
    $Q_\tau\leftarrow
    \operatorname{where}
    (M_\tau,Q_{\mathrm{alt}},Q_{\mathrm{rtn}})$

    \IF{$\|Q_\tau\neq Q_{\mathrm{rtn}}\|_0/|Q_\tau|\leq\eta$}
        \STATE
        $\mathcal{T}_{\mathrm{valid}}
        \leftarrow
        \mathcal{T}_{\mathrm{valid}}\cup\{\tau\}$
    \ENDIF
\ENDFOR

\FOR{each $\tau\in\mathcal{T}_{\mathrm{valid}}$}
    \STATE
    $W_\tau\leftarrow
    \operatorname{Dequant}(Q_\tau;\Delta,z)$
    \STATE
    $\displaystyle
    \mathcal{D}_{\mathrm{spec}}(\tau)
    \leftarrow
    \frac{
    \|\sigma_{1:k}(W_\tau)-\sigma_{1:k}(W)\|_2
    }{
    \|\sigma_{1:k}(W)\|_2+\epsilon
    }$
\ENDFOR

\STATE
$\displaystyle
\tau^\star\leftarrow
\arg\min_{\tau\in\mathcal{T}_{\mathrm{valid}}}
\mathcal{D}_{\mathrm{spec}}(\tau)$

\RETURN $Q^\star\leftarrow Q_{\tau^\star}$
\end{algorithmic}
\end{algorithm}

\section{Weight Reconstruction and ReRound Procedures}
\label{sec:supp_algorithms}

Algorithms~\ref{alg:weight_reconstruction} and
\ref{alg:reround} present the complete procedures for conditional
weight reconstruction and ReRound PTQ.
All quantization operations are applied group-wise, and all matrix
operations in Algorithm~\ref{alg:reround} are element-wise unless
otherwise stated.
The reconstructed matrices are produced once and reused for the W3 and
W4 ReRound runs.

Algorithm~\ref{alg:weight_reconstruction} reconstructs continuous
weights from 2-bit conditions using the conditioning parameters
$\Delta_c$ and $z_c$.
Algorithm~\ref{alg:reround} then maps $W$ and $W_{\mathrm{rec}}$ to the
same target W3 or W4 quantized space using $\Delta$ and $z$.


\bibliography{aaai2027}

@inproceedings{SahariaCSLWDGLKS22,
  author    = {Chitwan Saharia and
               William Chan and
               Saurabh Saxena and
               Lala Li and
               Jay Whang and
               Emily L. Denton and
               Kamyar Ghasemipour and
               Raphael Gontijo Lopes and
               Burcu Karagol Ayan and
               Tim Salimans and
               others},
  title     = {Photorealistic Text-to-Image Diffusion Models with Deep Language Understanding},
  booktitle = {NeurIPS},
  pages     = {36479--36494},
  year      = {2022},
}

@misc{DeepFloydTeam23,
  author       = {{DeepFloyd Team}},
  title        = {{DeepFloyd IF}: A Text-to-Image Diffusion Model},
  year         = {2023},
  howpublished = {\url{https://stability.ai/news-updates/deepfloyd-if-text-to-image-model}},
  note         = {Stability AI},
}

@inproceedings{HoJA20,
  author    = {Jonathan Ho and
               Ajay Jain and
               Pieter Abbeel},
  title     = {Denoising Diffusion Probabilistic Models},
  booktitle = {NeurIPS},
  pages     = {6840--6851},
  year      = {2020},
}

@article{SahariaHCSFN22,
  author  = {Chitwan Saharia and
             Jonathan Ho and
             William Chan and
             Tim Salimans and
             David J. Fleet and
             Mohammad Norouzi},
  title   = {Image Super-Resolution via Iterative Refinement},
  journal = {IEEE Transactions on Pattern Analysis and Machine Intelligence},
  pages   = {4713--4726},
  year    = {2022},
}

@article{Efron11,
  author  = {Bradley Efron},
  title   = {Tweedie's Formula and Selection Bias},
  journal = {Journal of the American Statistical Association},
  pages   = {1602--1614},
  year    = {2011},
}

@misc{Falcon3Team24,
  author = {{Falcon LLM Team}},
  title  = {The Falcon 3 Family of Open Models},
  year   = {2024},
  month  = {December},
  url    = {https://huggingface.co/blog/falcon3},
}

@article{GrattafioriDJKD24,
  author  = {Aaron Grattafiori and
             Abhimanyu Dubey and
             Abhinav Jauhri and
             Abhinav Pandey and
             Ahmad Al{-}Dahle and
             others},
  title   = {The Llama 3 Herd of Models},
  journal = {arXiv preprint arXiv:2407.21783},
  year    = {2024},
}

@inproceedings{BidermanSABHKPR23,
  author    = {Stella Biderman and
               Hailey Schoelkopf and
               Quentin Gregory Anthony and
               Herbie Bradley and
               Kyle O'Brien and
               Eric Hallahan and
               Mohammad Aflah Khan and
               Shivanshu Purohit and
               USVSN Sai Prashanth and
               Edward Raff and
               others},
  title     = {Pythia: A Suite for Analyzing Large Language Models Across Training and Scaling},
  booktitle = {ICML},
  pages      = {2397--2430},
  year       = {2023},
}

@misc{JavaheripiB23,
  author = {Mojan Javaheripi and
            S{\'e}bastien Bubeck},
  title  = {Phi-2: The Surprising Power of Small Language Models},
  year   = {2023},
  month  = {December},
  url    = {https://www.microsoft.com/en-us/research/blog/phi-2-the-surprising-power-of-small-language-models/},
}

@inproceedings{AllalLBMPTMKLS25,
  author    = {Loubna Ben Allal and
               Anton Lozhkov and
               Elie Bakouch and
               Gabriel Martin Bl{\'a}zquez and
               Guilherme Penedo and
               Lewis Tunstall and
               Andr{\'e}s Marafioti and
               Hynek Kydl{\'i}{\v{c}}ek and
               Agust{\'i}n Piqueres Lajar{\'i}n and
               Vaibhav Srivastav and
               others},
  title     = {SmolLM2: When Smol Goes Big---Data-Centric Training of a Fully Open Small Language Model},
  booktitle = {COLM},
  year      = {2025},
}

@article{TeamWSGLABGHJ25,
  author  = {{OLMo Team} and
             Pete Walsh and
             Luca Soldaini and
             Dirk Groeneveld and
             Kyle Lo and
             Shane Arora and
             Akshita Bhagia and
             Yuling Gu and
             Shengyi Huang and
             Matt Jordan and
             others},
  title   = {OLMo 2 Furious},
  journal = {arXiv preprint arXiv:2501.00656},
  year    = {2025},
}

@article{YangLAYZHBGYHL25,
  author  = {An Yang and
             Anfeng Li and
             Baosong Yang and
             Beichen Zhang and
             Binyuan Hui and
             Bo Zheng and
             Bowen Yu and
             Chang Gao and
             Chengen Huang and
             Chenxu Lv and
             others},
  title   = {Qwen3 Technical Report},
  journal = {arXiv preprint arXiv:2505.09388},
  year    = {2025},
}

@article{TeamKFPVMPRM25,
  author  = {{Gemma Team} and
             Aishwarya Kamath and
             Johan Ferret and
             Shreya Pathak and
             Nino Vieillard and
             Ramona Merhej and
             Sarah Perrin and
             Tatiana Matejovicova and
             Alexandre Ram{\'e} and
             Morgane Rivi{\`e}re and
             others},
  title   = {Gemma 3 Technical Report},
  journal = {arXiv preprint arXiv:2503.19786},
  year    = {2025},
}

@article{TeamRPSHBHMSRR24,
  author  = {{Gemma Team} and
             Morgane Riviere and
             Shreya Pathak and
             Pier Giuseppe Sessa and
             Cassidy Hardin and
             Surya Bhupatiraju and
             L{\'e}onard Hussenot and
             Thomas Mesnard and
             Bobak Shahriari and
             Alexandre Ram{\'e} and
             others},
  title   = {Gemma 2: Improving Open Language Models at a Practical Size},
  journal = {arXiv preprint arXiv:2408.00118},
  year    = {2024},
}

@inproceedings{AshkboosMCLCJAHH24,
  author    = {Saleh Ashkboos and
               Amirkeivan Mohtashami and
               Maximilian L. Croci and
               Bo Li and
               Pashmina Cameron and
               Martin Jaggi and
               Dan Alistarh and
               Torsten Hoefler and
               James Hensman},
  title     = {QuaRot: Outlier-Free 4-Bit Inference in Rotated LLMs},
  booktitle = {NeurIPS},
  pages     = {100213--100240},
  year      = {2024},
}

@inproceedings{DettmersPHZ23,
  author    = {Tim Dettmers and
               Artidoro Pagnoni and
               Ari Holtzman and
               Luke Zettlemoyer},
  title     = {QLoRA: Efficient Finetuning of Quantized LLMs},
  booktitle = {NeurIPS},
  pages     = {10088--10115},
  year      = {2023},
}

@inproceedings{MullerBZCBC26,
  author    = {Lorenz K. M{\"u}ller and
               Philippe Bich and
               Jiawei Zhuang and
               Ahmet {\c{C}}elik and
               Luca Benfenati and
               Lukas Cavigelli},
  title     = {SINQ: Sinkhorn-Normalized Quantization for Calibration-Free Low-Precision LLM Weights},
  booktitle = {ICML},
  year      = {2026},
}

@misc{BadriS23,
  author = {Hicham Badri and
            Appu Shaji},
  title  = {Half-Quadratic Quantization of Large Machine Learning Models},
  year   = {2023},
  month  = {November},
  url    = {https://dropbox.github.io/hqq_blog/},
}

@inproceedings{SunLBBZYYHJJ25,
  author    = {Yuxuan Sun and
               Ruikang Liu and
               Haoli Bai and
               Han Bao and
               Kang Zhao and
               Yuening Li and
               Xianzhi Yu and
               Lu Hou and
               Chun Yuan and
               Xin Jiang and
               others},
  title     = {FlatQuant: Flatness Matters for LLM Quantization},
  booktitle = {ICML},
  year      = {2025},
}

@inproceedings{LiuZFSCKCTB25,
  author    = {Zechun Liu and
               Changsheng Zhao and
               Igor Fedorov and
               Bilge Soran and
               Dhruv Choudhary and
               Raghuraman Krishnamoorthi and
               Vikas Chandra and
               Yuandong Tian and
               Tijmen Blankevoort},
  title     = {SpinQuant: LLM Quantization with Learned Rotations},
  booktitle = {ICLR},
  year      = {2025},
}

@inproceedings{XiaoLSWDH23,
  author    = {Guangxuan Xiao and
               Ji Lin and
               Mickael Seznec and
               Hao Wu and
               Julien Demouth and
               Song Han},
  title     = {SmoothQuant: Accurate and Efficient Post-Training Quantization for Large Language Models},
  booktitle = {ICML},
  pages      = {38087--38099},
  year       = {2023},
}

@inproceedings{ShaoCZXXLZPQL24,
  author    = {Wenqi Shao and
               Mengzhao Chen and
               Zhaoyang Zhang and
               Peng Xu and
               Lirui Zhao and
               Zhiqian Li and
               Kaipeng Zhang and
               Gao Peng and
               Yu Qiao and
               Ping Luo},
  title     = {OmniQuant: Omnidirectionally Calibrated Quantization for Large Language Models},
  booktitle = {ICLR},
  year      = {2024},
}

@article{LinTTYXH25,
  author  = {Ji Lin and
             Jiaming Tang and
             Haotian Tang and
             Shang Yang and
             Guangxuan Xiao and
             Song Han},
  title   = {AWQ: Activation-Aware Weight Quantization for On-Device LLM Compression and Acceleration},
  journal = {GetMobile: Mobile Computing and Communications},
  pages    = {12--17},
  year    = {2025},
}

@article{FrantarAHA22,
  author  = {Elias Frantar and
             Saleh Ashkboos and
             Torsten Hoefler and
             Dan Alistarh},
  title   = {GPTQ: Accurate Post-Training Quantization for Generative Pre-Trained Transformers},
  journal = {arXiv preprint arXiv:2210.17323},
  year    = {2022},
}

@article{SunBKTSDT25,
  author  = {Ziteng Sun and
             Adrian Benton and
             Samuel Kushnir and
             Asher Trockman and
             Vikas Singh and
             Suhas Diggavi and
             Ananda Theertha Suresh},
  title   = {CafeQ: Calibration-Free Quantization via Learned Transformations and Adaptive Rounding},
  journal = {arXiv preprint arXiv:2511.19705},
  year    = {2025},
}

@inproceedings{ChengZSCHLL24,
  author    = {Wenhua Cheng and
               Weiwei Zhang and
               Haihao Shen and
               Yiyang Cai and
               Xin He and
               Kaokao Lv and
               Yi Liu},
  title     = {Optimize Weight Rounding via Signed Gradient Descent for the Quantization of LLMs},
  booktitle = {Findings of EMNLP},
  pages     = {11332--11350},
  year      = {2024},
}

@article{ChengZGS25,
  author  = {Wenhua Cheng and
             Weiwei Zhang and
             Heng Guo and
             Haihao Shen},
  title   = {SignRoundV2: Closing the Performance Gap in Extremely Low-Bit Post-Training Quantization for LLMs},
  journal = {arXiv preprint arXiv:2512.04746},
  year    = {2025},
}

@inproceedings{LeeKKL23,
  author    = {Jung Hyun Lee and
               Jeonghoon Kim and
               Se Jung Kwon and
               Dongsoo Lee},
  title     = {FlexRound: Learnable Rounding Based on Element-Wise Division for Post-Training Quantization},
  booktitle = {ICML},
  pages     = {18913--18939},
  year      = {2023},
}

@inproceedings{NagelABLB20,
  author    = {Markus Nagel and
               Rana Ali Amjad and
               Mart van Baalen and
               Christos Louizos and
               Tijmen Blankevoort},
  title     = {Up or Down? Adaptive Rounding for Post-Training Quantization},
  booktitle = {ICML},
  pages     = {7197--7206},
  year      = {2020},
}

@inproceedings{DravidGWAWEA24,
  author    = {Amil Dravid and
               Yossi Gandelsman and
               Kuan-Chieh Wang and
               Rameen Abdal and
               Gordon Wetzstein and
               Alexei A. Efros and
               Kfir Aberman},
  title     = {Interpreting the Weight Space of Customized Diffusion Models},
  booktitle = {NeurIPS},
  pages     = {137334--137371},
  year      = {2024},
}

@inproceedings{SoroALJCHH25,
  author    = {Bedionita Soro and
               Bruno Andreis and
               Hayeon Lee and
               Wonyong Jeong and
               Song Chong and
               Frank Hutter and
               Sung Ju Hwang},
  title     = {Diffusion-based Neural Network Weights Generation},
  booktitle = {ICLR},
  year      = {2025},
}

@article{WangTZYXZZDLY24,
  author  = {Kai Wang and
             Dongwen Tang and
             Boya Zeng and
             Yida Yin and
             Zhaopan Xu and
             Yukun Zhou and
             Zelin Zang and
             Trevor Darrell and
             Zhuang Liu and
             Yang You},
  title   = {Neural Network Diffusion},
  journal = {arXiv preprint arXiv:2402.13144},
  year    = {2024},
}

@article{TangCXL25,
  author  = {Anqi Tang and
             Youming Chen and
             Shuchen Xue and
             Zhaoqiang Liu},
  title   = {Learning Single Index Models with Diffusion Priors},
  journal = {arXiv preprint arXiv:2505.21135},
  year    = {2025},
}

@article{MengK22,
  author  = {Xiangming Meng and
             Yoshiyuki Kabashima},
  title   = {Quantized Compressed Sensing with Score-Based Generative Models},
  journal = {arXiv preprint arXiv:2211.13006},
  year    = {2022},
}

@inproceedings{ChungKMKY23,
  author    = {Hyungjin Chung and
               Jeongsol Kim and
               Michael Thompson McCann and
               Marc Louis Klasky and
               Jong Chul Ye},
  title     = {Diffusion Posterior Sampling for General Noisy Inverse Problems},
  booktitle = {ICLR},
  year      = {2023},
}

@inproceedings{ChenL26,
  author    = {Youming Chen and
               Zhaoqiang Liu},
  title     = {Diffusion Model Based Signal Recovery Under 1-Bit Quantization},
  booktitle = {AAAI},
  pages     = {20400--20408},
  year      = {2026},
}

@inproceedings{VavilalaSF25,
  author       = {Vaibhav Vavilala and
                  Faaris Shaik and
                  David A. Forsyth},
  title        = {Dequantization and Color Transfer with Diffusion Models},
  booktitle    = {WACV},
  pages        = {9630--9639},
  year         = {2025},
}

@inproceedings{ClarkLCK0T19,
  author       = {Christopher Clark and
                  Kenton Lee and
                  Ming{-}Wei Chang and
                  Tom Kwiatkowski and
                  Michael Collins and
                  Kristina Toutanova},
  title        = {BoolQ: Exploring the Surprising Difficulty of Natural Yes/No Questions},
  booktitle    = {NAACL-HLT},
  pages        = {2924--2936},
  year         = {2019},
}

@inproceedings{SapRCBC19,
  author       = {Maarten Sap and
                  Hannah Rashkin and
                  Derek Chen and
                  Ronan Le Bras and
                  Yejin Choi},
  title        = {Social IQa: Commonsense Reasoning about Social Interactions},
  booktitle    = {EMNLP-IJCNLP},
  pages        = {4462--4472},
  year         = {2019},
}

@inproceedings{BiskZLGC20,
  author       = {Yonatan Bisk and
                  Rowan Zellers and
                  Ronan Le Bras and
                  Jianfeng Gao and
                  Yejin Choi},
  title        = {{PIQA:} Reasoning about Physical Commonsense in Natural Language},
  booktitle    = {AAAI},
  pages        = {7432--7439},
  year         = {2020},
}

@inproceedings{SakaguchiBBC20,
  author       = {Keisuke Sakaguchi and
                  Ronan Le Bras and
                  Chandra Bhagavatula and
                  Yejin Choi},
  title        = {WinoGrande: An Adversarial Winograd Schema Challenge at Scale},
  booktitle    = {AAAI},
  pages        = {8732--8740},
  year         = {2020},
}

@inproceedings{MerityX0S17,
  author       = {Stephen Merity and
                  Caiming Xiong and
                  James Bradbury and
                  Richard Socher},
  title        = {Pointer Sentinel Mixture Models},
  booktitle    = {ICLR},
  year         = {2017},
}

@article{RaffelSRLNMZLL20,
  author       = {Colin Raffel and
                  Noam Shazeer and
                  Adam Roberts and
                  Katherine Lee and
                  Sharan Narang and
                  Michael Matena and
                  Yanqi Zhou and
                  Wei Li and
                  Peter J. Liu},
  title        = {Exploring the Limits of Transfer Learning with a Unified Text-to-Text
                  Transformer},
  journal      = {Journal of machine learning research},
  volume       = {21},
  pages        = {140:1--140:67},
  year         = {2020},
}

@misc{eval-harness,
  author       = {Leo Gao and
                  Jonathan Tow and
                  Baber Abbasi and
                  Stella Biderman and
                  Sid Black and
                  Anthony DiPofi and
                  Charles Foster and
                  Laurence Golding and
                  Jeffrey Hsu and
                  Alain Le Noac'h and
                  Haonan Li and
                  Kyle McDonell and
                  Niklas Muennighoff and
                  Chris Ociepa and
                  Jason Phang and
                  Laria Reynolds and
                  Hailey Schoelkopf and
                  Aviya Skowron and
                  Lintang Sutawika and
                  Eric Tang and
                  Anish Thite and
                  Ben Wang and
                  Kevin Wang and
                  Andy Zou},
  title        = {The Language Model Evaluation Harness},
  year         = {2026},
  howpublished = {Zenodo},
  note         = {Version v0.4.11, released February 2026}
}


\end{document}